\documentclass[pdflatex,sn-mathphys-num]{sn-jnl}

\usepackage{epsfig}
\usepackage[inkscapelatex=false]{svg}
\usepackage{graphicx}%
\usepackage{multirow}%
\usepackage{multicol}
\usepackage{amsmath,amssymb,amsfonts}%

\usepackage{mathrsfs}%
\usepackage[title]{appendix}%
\usepackage{textcomp}%

\usepackage{booktabs}%
\usepackage{algorithm}%
\usepackage{algorithmicx}%
\usepackage{algpseudocode}%
\usepackage{listings}%

\newcommand{\std}[1]{\textsubscript{\tiny{$\pm$#1}}}

\usepackage[normalem]{ulem}
\useunder{\uline}{\ul}{}
\usepackage{xspace}
\usepackage[table, dvipsnames]{xcolor}
\usepackage{array}

\newcommand{\mName}{SPCL\xspace}
\usepackage{anyfontsize}

\begin{document}

\title[Article Title]{Leveraging Self-Paced Curriculum Learning for Enhanced Modality Balance in Multimodal Conversational Emotion Recognition}



\author{\fnm{Phuong-Anh} \spfx{Nguyen}}\email{22028332@vnu.edu.vn}

\author{\fnm{The-Son} \spfx{Le}}\email{21020089@vnu.edu.vn}

\author{\fnm{Duc-Trong} \spfx{Le}}\email{trongld@vnu.edu.vn}

\author*{\fnm{Cam-Van Thi} \spfx{Nguyen*}}\email{vanntc@vnu.edu.vn}

\affil{\orgdiv{Faculty of Information Technology,  } \orgname{VNU University of Engineering and Technology}, \orgaddress{\postcode{100000},  \city{Hanoi}, \country{Vietnam}}}


\abstract{Multimodal Emotion Recognition in Conversations (MERC) is a crucial task for understanding human interactions, where multimodal approaches integrating language, facial expressions, and vocal tone have led to significant advancements. However, a persistent challenge in multimodal architectures, including MERC, is modality misalignment and imbalanced learning. These issues often hinder models from effectively utilizing multimodal information, leading to suboptimal performance despite the availability of multiple modalities.
To address this, we design a framework for MERC with a proposed plug-and-play module that builds upon Self-Paced Curriculum Learning (\textbf{SPCL}). As in Curriculum Learning (CL), an effective \textit{Difficulty Measurer} is essential for structuring a meaningful \textit{Learning Scheduler}. In this work, we propose a dual-level Difficulty Measurer tailored for MERC, addressing both intra- and inter-conversational dynamics.
Unlike conventional approaches that assess difficulty only at the utterance level, our dual-level design incorporates a conversation-level difficulty score. The utterance-level score captures fine-grained modality-specific challenges, while the conversation-level score models broader dialogue structures, including emotional dependencies and modality coherence within the conversation. This holistic evaluation enables our Learning Scheduler to dynamically guide training from easier to more challenging instances.
By integrating SPCL into existing MERC architectures, our method effectively mitigates modality imbalance and enhances model robustness. Extensive experiments on the IEMOCAP and MELD datasets confirm consistent improvements: on IEMOCAP, SPCL achieves gains ranging from approximately +1.2\% to +6.6\% in weighted F1-score over baseline models across different architectures and modality settings, while on MELD, it delivers even more pronounced improvements, with gains reaching up to +10.4\% over baseline models. These gains underscore the practical value of SPCL for real-world MERC applications, as it substantially improves emotion recognition accuracy while maintaining compatibility as a plug-and-play module across diverse model architectures.}




\maketitle

\section{Introduction}
\label{sec:intro}
Multimodal learning has gained significant attention in artificial intelligence due to its ability to integrate information from diverse sources, such as text (T), audio (A), and visual (V) data \cite{yuan2025survey, baltruvsaitis2018multimodal}. By leveraging complementary modalities, multimodal models enhance understanding and improve predictive performance~\cite{liang2021multibench, liang2024foundations}. Among its diverse applications, Emotion Recognition in Conversations (ERC) has emerged as a critical task spanning artificial intelligence, cognitive science, and social sciences. The primary objective of ERC is to detect the emotional undertones accompanying each utterance in a conversation. While language plays a crucial role, emotions are often conveyed through a combination of verbal and non-verbal cues, such as facial expressions, tonal variations, and bodily gestures~\cite{gladys2023survey}. Given the inherently multimodal nature of human communication, integrating multimodal data into ERC has naturally evolved into Multimodal Emotion Recognition in Conversations (MERC). By jointly modeling text, audio, and visual modalities, MERC seeks to enhance emotion recognition beyond the constraints of unimodal approaches~\cite{ghosal-etal-2020-cosmic, hu2021mmgcn, nguyen-etal-2023-conversation}. 

However, the effectiveness of multimodal integration is often limited by modality imbalance, where certain modalities contribute disproportionately to learning, leading to suboptimal model performance. Prior research has examined this phenomenon from various perspectives, describing it as the dominance of specific modalities~\cite{peng2022balanced}, discrepancies in convergence rates~\cite{wang2020makes, shi2025gradient}, or diminishing marginal utility of modalities~\cite{wang2023unlocking}. \citet{zhang2024multimodal} categorize these challenges into Property Discrepancy and Quality Discrepancy. Property Discrepancy arises because different modalities exhibit distinct learning behaviors due to their heterogeneous nature. For instance, audio features often require fewer training iterations before overfitting, whereas visual features typically converge more slowly. This inconsistency complicates optimization, making it difficult to balance learning across modalities. Quality Discrepancy refers to the uneven distribution of task-relevant information among modalities. While all modalities aim to represent the same underlying context, some provide stronger discriminative signals than others. Multimodal models, inherently greedy in their learning dynamics, tend to prioritize the most informative modality while underutilizing weaker ones, further exacerbating modality imbalance.

\begin{figure}[!ht]
    \centering
    \includegraphics[width=0.7\textwidth]{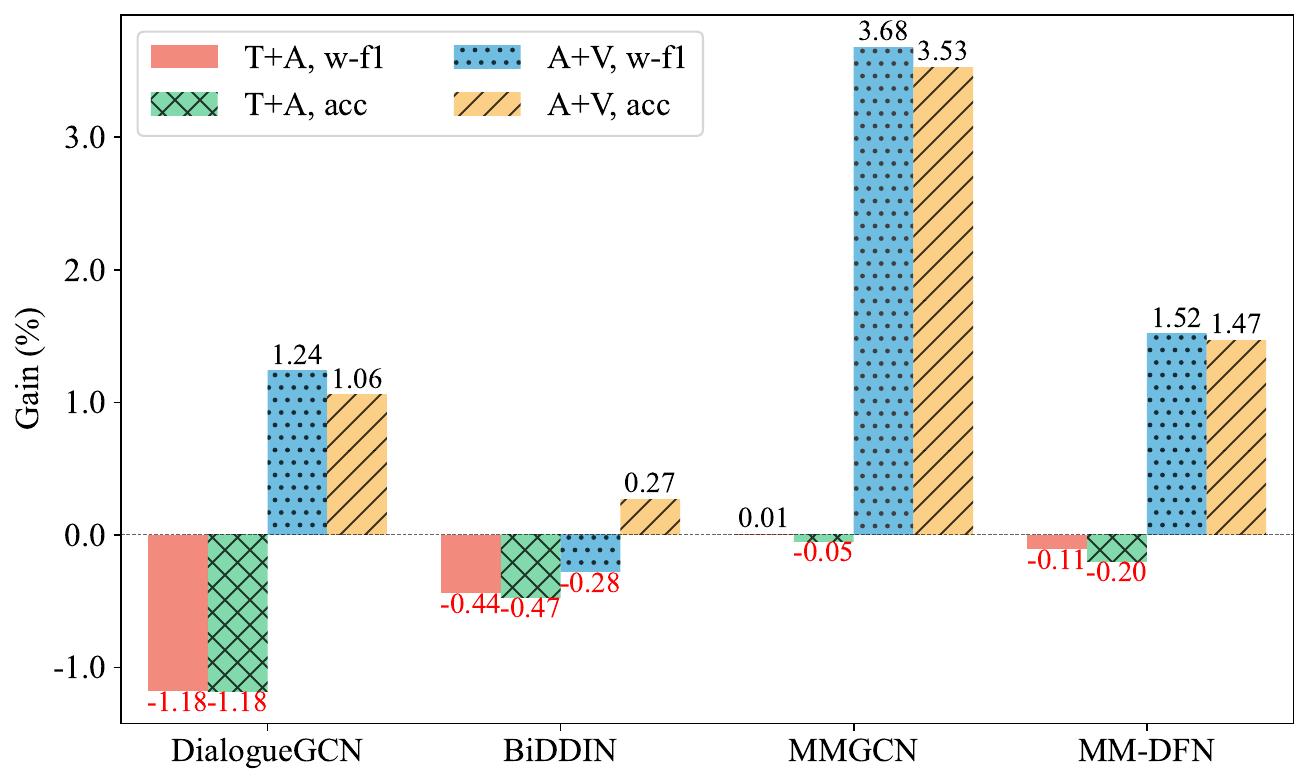}
    \caption{Performance gain of training baseline models on full modality (ATV) over training on bi-modality (AT or TV). Evaluation uses W-F1 and Acc metrics across IEMOCAP dataset.}
    \label{fig:gain-iemocap}
\end{figure} 

In MERC, text is often the dominant modality, providing explicit emotional cues, whereas audio and visual features are more subtle and context-dependent. Consequently, models may become overly reliant on textual data, failing to fully integrate information from other modalities. This imbalance ultimately leads to suboptimal fusion strategies and reduced overall performance. As shown in Figure \ref{fig:gain-iemocap}, we conducted an experiment comparing the performance of tri-modal models (ATV) with their bi-modal versions (TA and AV). While there were instances where the tri-modal approach outperformed its bi-modal counterparts, the results were inconsistent. In half of the cases, incorporating all three modalities led to a performance reduction, with accuracy dropping between 0.05\% and 1.18\%, and w-F1 dropping between 0.11\% and 1.18\%. These findings highlight the need to further investigate modality imbalance, a crucial factor affecting MER performance.
To address modality imbalance in MERC, previous studies have explored several strategies, including pre-trained unimodal networks~\cite{du2023uni, wu2022characterizing}, auxiliary learning objectives~\cite{xu2023mmcosine, zhou2023intra}, and optimization-based techniques~\cite{peng2022balanced, fan2023pmr, nguyenrobult}. Pre-trained unimodal networks leverage separately trained feature extractors for each modality before integrating them into a multimodal framework, improving individual modality representations but often incurring high computational costs and requiring large-scale labeled data~\cite{ismail2020improving, huang2025adaptive}. Auxiliary learning objectives introduce additional constraints or tasks, such as contrastive or self-supervised learning, to enhance modality-specific representations~\cite{zhou2023intra, wulearning}, yet they may struggle to generalize across diverse conversational contexts. Optimization-based methods, which directly adjust gradient updates to prevent modality dominance, have shown promise in balancing multimodal contributions~\cite{wang2023unlocking, peng2022balanced, li2023boosting, liu2025reward}, but their reliance on complex gradient manipulation makes them difficult to implement and tune.

Building upon these persistent challenges, we propose a novel approach based on Self-Paced Curriculum Learning to address modality imbalance in MERC. As an extension of Curriculum Learning, Self-Paced Curriculum Learning enhances this process by dynamically selecting training samples based on the model’s learning progress, ensuring a more adaptive and balanced training trajectory. 
Specifically, we introduce a framework specifically designed for MERC, incorporating our proposed \textbf{SPCL} module to tackle modality imbalance. Our proposed SPCL module comprises two key components: (1) a \textbf{Difficulty Measurer}, which assesses the complexity of each training sample through a dual-level design that captures utterance-level recognition performance and conversation-level modality discrepancy to ensure balanced learning across modalities; and (2) a \textbf{Learning Scheduler}, which dynamically refines sample selection based on the model’s evolving competence, progressively guiding training from easier to more challenging instances. 

While modality imbalance has been explored in general multimodal learning, its unique challenges in MERC, especially those linked to conversational context and dialogue structures, have not been fully addressed. Our method fills this gap by offering an SPCL strategy that is carefully tailored for MERC. With this design, \mName{} improves model stability, reduces problems from modality imbalance, and achieves better performance on MERC tasks.

Overall, our key contributions are as follows:
\begin{itemize}
\item We introduce a novel framework tailored for MERC, integrating our proposed module, \mName{}, based on Self-Paced Curriculum Learning (SPCL), to address modality imbalance through adaptive sample selection.
\item We design a Difficulty Measurer with a dual-level assessment mechanism, capturing both utterance-level recognition performance and conversation-level modality discrepancy. This ensures a more balanced multimodal representation learning process. Additionally, we design a Learning Scheduler that dynamically selects training samples based on model competence, facilitating a progressive learning process that enhances model robustness.
\item We conduct extensive experiments on benchmark MERC datasets, IEMOCAP~\cite{busso2008iemocap} and MELD~\cite{poria2019meld}, integrating SPCL as a plug-in module into four baseline models. Results demonstrate significant performance improvements, validating the effectiveness of our approach.
\end{itemize}

The remainder of this paper is organized as follows: Section \ref{sec:related} provides a comprehensive review of related work, highlighting existing approaches to modality imbalance in MERC. Section \ref{sec:method} introduces our proposed method, detailing the architecture and functionality of the SPCL module. Section \ref{sec:exp-setting} describes the experimental setup, including datasets, baseline models, and evaluation metrics. Section \ref{sec:results} presents our results, with an in-depth analysis of performance improvements, ablation studies, and discussions on the impact of each component. Finally, Section \ref{sec:conclude} concludes the paper and outlines potential directions for future research.
 
\section{Related Work}
\label{sec:related}
\subsection{Multimodal Conversational Emotion Recognition}
Multimodal Emotion Recognition in Conversation (MERC) aims to identify emotions by leveraging multiple modalities, including textual, auditory, and visual data. The interplay between these modalities enhances emotion recognition, yet the complexity of conversational dynamics—such as speaker interactions and evolving emotional states—poses significant challenges. To tackle these issues, various methodologies have been developed, ranging from sequential modeling to advanced fusion and graph-based approaches~\cite{hu2021mmgcn, ghosal2019dialoguegcn, nguyen-etal-2023-conversation, nguyen-etal-2024-curriculum}. 

Early MERC models primarily relied on recurrent neural networks (RNNs) and transformers to model dialogue context. DialogueGCN~\cite{ghosal2019dialoguegcn} structures conversations as graphs, capturing speaker relationships through edge connections. COSMIC~\cite{ghosal-etal-2020-cosmic} integrates commonsense knowledge into an RNN-based framework to enhance context understanding, while DialogueCRN~\cite{hu-etal-2021-dialoguecrn} employs cognitive-inspired mechanisms to model emotion flow across utterances. More recently, DialogXL~\cite{shen2021dialogxl} has demonstrated the effectiveness of transformers in processing long-range dependencies within conversations.
A critical aspect of MERC is how modalities are integrated. Initial approaches, such as bc-LSTM~\cite{poria-etal-2017-context} and CMN~\cite{hazarika-etal-2018-conversational}, focused on concatenating modality features but lacked sophisticated interaction mechanisms. Subsequent methods have improved fusion techniques by leveraging hierarchical structures~\cite{hazarika-etal-2018-icon}, feature disentanglement~\cite{li2023revisiting}, and attention-based co-learning~\cite{shi-huang-2023-multiemo}. Transformer-based models like TBJE~\cite{delbrouck-etal-2020-transformer} apply modular co-attention to refine cross-modal representations, while AdaIGN~\cite{tu2024adaptive} selectively adjusts node and edge relationships within the fused feature space.
Beyond sequential and fusion-based models, graph neural networks (GNNs) have emerged as a powerful tool for encoding conversational structures. MMGCN~\cite{hu2021mmgcn} captures multimodal dependencies through graph convolution, considering both speaker identity and context flow. COGMEN~\cite{joshi2022cogmen} extends this by incorporating conversational graphs that dynamically evolve over time. Meanwhile, CORECT~\cite{nguyen-etal-2023-conversation} models relational interactions between utterances, allowing for refined representation learning across dialogues.

Despite their strengths, these methods primarily focus on architectural innovations and often overlook the critical issue of modality imbalance. Addressing this challenge requires not only optimizing fusion strategies but also devising techniques to balance the contributions of different modalities, ensuring robust performance even when certain modalities dominate or underperform.

\subsection{Imbalanced Multimodal Learning}
Multimodal learning, a rapidly growing field in artificial intelligence, focuses on leveraging and integrating data from multiple modalities, such as text, images, and audio, to improve model performance and enable richer understanding~\cite{baltruvsaitis2018multimodal}. A critical challenge in multimodal learning is effectively integrating information from different modalities to enable complementary interactions. Traditional fusion strategies, such as early fusion, intermediate fusion, and late fusion, aim to combine multimodal information effectively. However, these strategies exhibit limited ability to resolve modality competition~\cite{huang2022modality} and imbalanced multimodal learning, where dominant modalities overshadow others. This imbalance often results in modality inhibition~\cite{fan2024detached, wei2025diagnosing}, where weaker modalities fail to contribute meaningfully to the final decision. As a consequence, when a dominant modality is missing or corrupted~\cite{guo2024multimodal}, the overall performance of these models often degrades significantly. Addressing this challenge requires balancing the contributions of all modalities while ensuring that weaker modalities are not entirely suppressed.

Recent studies~\cite{peng2022balanced, huang2022modality, nguyen2024ada2i} highlight that multimodal models often fail to outperform their best unimodal counterparts due to modality imbalance. To tackle this, various works have introduced optimization strategies aimed at rebalancing the learning process across modalities. One common approach is gradient modulation, where the model dynamically adjusts learning rates or gradients based on modality importance. For instance, \citet{peng2022balanced} propose an on-the-fly gradient modulation strategy that monitors the contribution discrepancies of each modality toward the learning objective, ensuring balanced optimization. Similarly, \citet{fan2023pmr} introduce a prototypical modal rebalance (PMR) method that controls updating directions for each modality, allowing for more effective unimodal learning. FAGM~\cite{wang2023unlocking} extends this by fine-tuning gradient updates at the parameter level, proportionally adjusting contributions from each modality to prevent over-reliance on dominant features.

Beyond gradient-based solutions, some methods enhance modality interaction to mitigate imbalance. RNA loss~\cite{planamente2022domain} introduces constraints in the loss function to align feature norms across modalities, ensuring more consistent feature representations and reducing modality discrepancies. OGM-GE~\cite{peng2022balanced} further improves balancing by dynamically adjusting gradients based on each modality’s importance, preventing weaker modalities from being overshadowed. Other approaches focus on modality interaction enhancements. MLA~\cite{wu2022characterizing} employs an alternating learning algorithm that iteratively updates different modalities, promoting stronger cross-modal dependencies. ReconBoost~\cite{hua2024reconboost} leverages gradient boosting to dynamically adjust learning objectives, capturing underutilized information from weaker modalities. Knowledge distillation has also been explored as a way to improve weaker modalities, with UMT~\cite{du2021improving} transferring knowledge from well-trained unimodal teachers to guide multimodal representations. More recently, On-the-fly Prediction Modulation (OPM)~\cite{wei2024fly} has been proposed as another approach to address modality imbalance. By monitoring the discriminative discrepancy between modalities during training, OPM dynamically drops features from the dominant modality with a certain probability, whereas OGM-GE~\cite{peng2022balanced} instead mitigates gradient contributions on-the-fly to balance learning across modalities. 
While they have demonstrated effectiveness in general multimodal tasks, they often introduce increased model complexity and additional training overhead.

Despite their contributions, these methods are often constrained by assumptions about network architecture, loss functions, or optimization methods, limiting their applicability in more general scenarios. Unlike these approaches, our method removes these restrictions by supporting arbitrary numbers of modalities, optimizers, and loss functions.
Furthermore, our method is specifically designed for the MERC task under imbalanced scenarios, addressing both utterance-level and conversation-level modality imbalance. We not only consider the external disparity across modalities but also explore the intrinsic factors within conversations that contribute to imbalance. 

\subsection{Self-paced Curriculum Learning}
Inspired by the structured progression of knowledge acquisition in human cognition, Curriculum Learning (CL)~\cite{bengio2009curriculum, soviany2022curriculum} has emerged as a training paradigm that organizes the learning process by introducing samples in an incremental order of complexity—from simpler to more challenging examples. By steering the model toward an optimal parameter space, CL has demonstrated significant potential across diverse domains, including large language models~\cite{wang2024curriculum}, action recognition~\cite{tong2022semi}, affective computing~\cite{nguyen-etal-2024-curriculum, yu2025tacl}, and reinforcement learning~\cite{narvekar2020curriculum}. A typical curriculum framework is composed of two fundamental components: a difficulty measurer, which quantifies the complexity of training samples, and a scheduler, which determines the timing and strategy for incorporating more complex samples into the training process. 
For the MERC task, \citet{nguyen-etal-2024-curriculum} employ a Directed Acyclic Graph to integrate textual, acoustic, and visual features within a unified framework. Their model leverages CL to address challenges related to emotional shifts and data imbalance; however, it does not specifically focus on modality imbalance.
In the broader context of multimodal imbalance learning, \citet{qian2025dyncim} propose a sample-level curriculum that dynamically assesses each sample’s difficulty based on prediction deviation, consistency, and stability. They also introduce a modality-level curriculum to measure modality contributions from both global and local perspectives. Nevertheless, their method does not directly address the unique challenges posed by MERC tasks.

A key subset of CL, known as Self-Paced Learning (SPL), automates difficulty evaluation by using the model’s current training loss as an indicator of sample complexity. Inspired by educational methodologies—where learners control their study pace by selecting topics, determining study methods, and managing learning duration~\cite{tullis2011effectiveness, han2025climd, zhou2025sample}—SPL offers a dynamic and adaptive training process.
Unlike traditional curriculum learning, which follows predefined criteria for structuring the learning process, SPL dynamically adjusts the training curriculum based on the model’s learning progress. By leveraging a loss-driven difficulty measurer, SPL adapts to different tasks and data distributions, ensuring a more tailored and efficient training process. Moreover, SPL seamlessly integrates curriculum design into the learning objective, making it a flexible tool for enhancing various machine learning frameworks~\cite{wang2021survey}.

In this work, we apply SPL to multimodal learning under imbalanced scenarios, particularly in MERC task. Our approach addresses modality discrepancy problem by introducing a Learning Scheduler strategy that designs an adaptive curriculum, dynamically selecting appropriate samples at each training step based on the model’s response. This ensures a balanced learning process, enabling the model to effectively handle modality imbalance while improving overall robustness.

\section{Methodology}
\label{sec:method}
\begin{figure*}
    \centering
    \includegraphics[width=\linewidth]{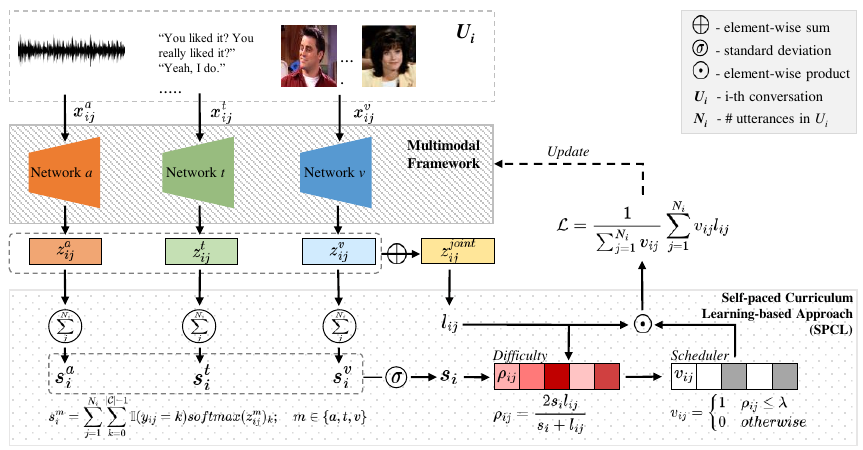}
    \caption{Our framework pipeline with integrated SPCL module.}
    \label{fig:model}
\end{figure*}

In this section, we introduce the detailed architecture of our framework with our proposed Self-Paced Curriculum Learning-based (SPCL) module, designed to mitigate modality imbalance in MERC.
Figure \ref{fig:model} illustrates the overall pipeline of our approach, showcasing how the SPCL module seamlessly integrates with existing MER models. In the following subsections, we provide an end-to-end overview of our framework, from obtaining unimodal emotion predictions to formulating the learning objective.We provide a detailed analysis of SPCL, including the formulation, implementation, and influence of the Difficulty Measurer and Learning Scheduler on the training process.

\subsection{Problem Definition} 
Given a predefined emotion category set $\mathcal{C}$ and a dataset $\mathcal{D}=\{U_1, U_2,\dots,U_{|\mathcal{D}|}\}$, conversation $U_i$ consists of $N_i$ utterances and their corresponding labels $\{(x_{i1},y_{i1}), (x_{i2}, y_{i2}),\dots, (x_{iN_i}, y_{iN_i})\}$. The \textit{Emotion Recognition in Conversation (ERC)} task is to predict an emotion label from $\mathcal{C}$ for each utterance $x_{ij}$. In the context of multimodal conversation, every utterance is represented through $M$ modalities. Specifically, the modalities include audio ($a$), textual ($t$) and visual ($v$) modal. Thus, the input can be written as: 
\begin{equation*}
    x_{ij} = \{x^a_{ij}, x^t_{ij}, x^v_{ij}\}
\end{equation*}
where $x_{ij}^m\in\mathbb{R}^{d_m}$ with $d_m$ is the dimension the $m$ modality.

In the following subsections, we introduce our framework, which includes 2 main sub-modules: \textit{(1) Modality Prediction, (2) Self-paced Curriculum Learning-based  (SPCL) module}.

\subsection{Modality Prediction}
\label{sec:modality-pred}
For each utterance $x_{ij}$ in conversation $U_i$, an emotion prediction network is utilized to generate uni-modal prediction logit:
\begin{equation}
    z_{ij}^m = \phi_m(x_{ij}^m; \theta^m), m\in\{a, t, v\}
    \label{eq:uni-pred}
\end{equation}
where the function $\phi_m(\cdot): \mathbb{R}^{d_m}\to \mathbb{R}^{|C|}$ is the uni-modal network with learnable parameter $\theta^m$, and $z_{ij}^m\in\mathbb{R}^{|\mathcal{C}|}$ is the logit prediction corresponding to modal m.

In order to retrieve the ultimate prediction for utterance $x_{ij}$, i.e. cross-modal logit, we perform a simple fusion step on the uni-modal logits from above. Specifically, the cross-modal logit is the sum of uni-modal logits. This fusion step can be written as: 
\begin{equation}
    z_{ij}^{joint} = \sum_{m}^{\{a,t,v\}}z_{ij}^m
    \label{eq:cross-pred}
\end{equation}
where $z_{ij}^{joint}\in\mathbb{R}^{ |\mathcal{C}|}$ is the cross-modal logit prediction for utterance $x_{ij}$, and $z_{ij}^{m}$ are the uni-modal logits from (Equation~\ref{eq:uni-pred}).

We evaluate the model's overall performance on utterance $x_{ij}$ via the cross-modal logit as follow:
\begin{equation}
    l_{ij} = -\log (softmax(z_{ij}^{joint})_{y_{ij}})
    \label{eq:utt-loss}
\end{equation}
where $l_{ij}\in \mathbb{R}$ is the loss with regard to utterance $x_{ij}$. 
\subsection{Self-paced Curriculum Learning-based Approach (SPCL)}
Self-paced Curriculum Learning is employed to alleviate the imbalance between modals during training; at the same time, elevate MERC task performance. Adapting existing works using curriculum learning, we also design the learning curricula via two components: (1) \textit{Difficulty Measurer} to determine the difficulty of all samples in the dataset, and (2) \textit{Learning Scheduler} to control the learning pace. 
\subsubsection*{Difficulty Measurer}
Traditional Difficulty Measurers treat each utterance as an independent sample, computing an utterance-level difficulty score to guide the sample selection process. In line with this approach, we adopt the utterance loss from Section \ref{sec:modality-pred} as the \textit{utterance-level score}, under the assumption that a higher loss indicates greater misalignment among the unimodal representations of the utterance. However, this method is limited by its inability to capture the broader dialogue dynamics. Specifically, an utterance-level score alone fails to capture broader conversational characteristics, such as emotional dependencies between utterances and the overall emotion distribution within a dialogue.

To address this limitation, we propose incorporating a \textit{conversation-level} difficulty score alongside the utterance-level score to provide a more comprehensive assessment of sample difficulty. This conversation-level score, which is unique to each dialogue, is shared among all utterances within the same conversation, thereby ensuring a more holistic evaluation of multimodal interactions and emotional coherence. 

To obtain the conversation-level score, firstly, we retrieve uni-modal scores for conversation $U_i$ using uni-modal logits as follow:
\begin{equation}
    s_i^m = \sum_{j=1}^{N_i}\sum_{k=0}^{|\mathcal{C}|-1}\mathbb{I}{(y_{ij}=k)}softmax(z_{ij}^m)_k
    \label{eq:uni-score}
\end{equation}
where $s_i^m$ is a scalar that acts as the score of $U_i$ w.r.t modal $m$, $z_{ij}^m$ is the logit from Equation~\ref{eq:uni-pred}, $\mathbb{I}{(y_{ij}=k)}$ is the indicator which equals 1 if $y_{ij}=k$ and 0 if otherwise, $softmax(\cdot)_k$ indicates the $k^{th}$ value of softmax.

Next, we derive the cross-modal score of \( U_i \) by computing the standard deviation of uni-modal scores, which serves as a quantifiable measure of inter-modal variation. The standard deviation function evaluates the extent to which individual uni-modal scores \( s_i^m \) deviate from their mean, thereby offering a systematic assessment of modality misalignment. A higher standard deviation signifies pronounced discrepancies among modalities, indicating that certain modalities exert greater influence while others contribute minimally, ultimately leading to an imbalanced representation. Conversely, a lower standard deviation suggests a more equitable contribution across modalities, facilitating more coherent multimodal information fusion. Furthermore, standard deviation is a robust statistical measure that normalizes variations across different datasets, ensuring a consistent and reliable evaluation of modality divergence. Consequently, we define this function’s output as the \textit{conversation-level score}, leveraging its capability to effectively capture and quantify modality misalignment. Formally, this score is computed as follows:
\begin{equation}
    s_i = \sigma(s^a_i, s^t_i, s^v_i)
    \label{eq:conv-score}
\end{equation}
where $s_i$ is the conversation-level score of $U_i$, $s_i^m$ with $m\in\{a,t,v\}$ are from Equation~\ref{eq:uni-score}, and $\sigma(\cdot)$ is the standard deviation function.

Finally, the final \textit{difficulty} of utterance $x_{ij}$ is formulated from utterance-level score $l_{ij}$ and conversation-level score $s_i$. To ensure that the difficulty fairly represents both ER task and modality discrepancy, we combine these two values the using harmonic mean:
\begin{equation}
    \rho_{ij} = \frac{2 s_i l_{ij}}{s_i + l_{ij}}
    \label{eq:diffc}
\end{equation}
where $\rho_{ij} \in \mathbb{R}$ is the difficulty of utterance $x_{ij}$. The use of the harmonic mean is particularly advantageous in this context, as it penalizes extreme values and ensures that neither $l_{ij}$ (recognition difficulty) nor $s_i$ (modality misalignment) dominates the difficulty calculation. Unlike the arithmetic mean, which can be disproportionately influenced by large values, the harmonic mean emphasizes situations where both components are relatively balanced, prevents overly sensitive to one factor while ignoring the other. This formulation adaptively scales difficulty, allowing the SPCL to prioritize utterances that are easy to classify or exhibit insignificant modality misalignment, i.e. utterances with low $\rho_{ij}$.

\subsubsection*{Learning Scheduler}

A learning scheduler, responsible for organizing and distributing training samples throughout the learning process, is employed to ensure such structured training. Here, we adopt a controlled approach by utilizing a hard regularizer, ensuring a strictly progressive training schedule where training samples are either included or excluded based on their predefined difficulty level $\rho_{ij}$. Unlike soft regularization techniques, which gradually adjust sample importance through continuous weighting~\cite{hacohen2019power}, a hard regularizer completely excludes difficult samples until the model is sufficiently trained to handle them. This strict progression helps prevent catastrophic forgetting and encourages a more stable knowledge accumulation process, as demonstrated in prior curriculum learning research~\cite{soviany2022curriculum}.

Specifically, we define a mask value $v_{ij}$ corresponding to utterance $x_{ij}$. Our $v_{ij}$ is retrieved using a \textit{hard regularizer} $g(\rho_{ij}, \lambda)$ that leads to a binary weighting:
\begin{equation}
    v_{ij} = g(\rho_{ij},\lambda) =
    \begin{cases}
        1  &\rho_{ij} \leq \lambda, \\
        0  &otherwise
    \end{cases}
    \label{eq:mask}
\end{equation}
where $\rho_{ij}$ is the difficulty, and $\lambda >0$ is a threshold parameter that acts as the boundary splitting easy and hard samples.

To ensure adherence to the intended learning manner, the threshold parameter $\lambda$ is initially set to a relatively small value. As training progresses, $\lambda$ is gradually increased to regulate the difficulty level of the samples introduced to the model. Specifically:
\begin{equation}
    \lambda^{(t)} = \begin{cases}
       \varepsilon &t=0, \\
        \alpha\lambda^{(t-1)}  &t>0
    \end{cases}
    \label{eq:lambda-upd}
\end{equation}
where $\lambda^{(t)}$ is the difficulty threshold at epoch $t-th$, $\varepsilon$ is a relatively small number, $\alpha>1$ is the aging hyper-parameter used to monitor the learning pace. Here,  $\varepsilon$ and $\alpha$ are hand-selected via conducting experiments.  
\subsection{Multi-modal Learning with SPCL}

In a standard MERC task, the objective loss is computed as the sum of the negative log likelihood loss for each sample $l_{ij}$, as follows:
\begin{equation}
    \mathcal{L} = \frac{1}{\sum_{i=1}^{|\mathcal{D}|}\sum_{j=1}^{N_i}1} \sum_{i=1}^{|\mathcal{D}|}\sum_{j=1}^{N_i}l_{ij}
    \label{eq:init-loss}
\end{equation}
with $\sum_{i=1}^{|\mathcal{D}|}\sum_{j=1}^{N_i}1$ refers to the total number of utterances in the dataset.

At each training step, once the loss is obtained, the parameters of the multimodal framework, i.e. the learnable parameters $\theta^m$ of the uni-modal networks $\phi^m(\cdot)$, are updated using gradient-based optimization as following:
\begin{equation}
    \theta^{m^{(t+1)}} \leftarrow \theta^{m^{(t)}} - \eta\frac{\partial \mathcal{L}}{\partial \theta^{m^{(t)}}}, \quad m\in\{a, t, v\}
\end{equation}

However, with the integration of our SPCL module, we refine this process by selectively excluding difficult samples from the loss computation. Specifically, we mitigate the influence of these challenging samples by scaling \( l_{ij} \) with a binary mask \( v_{ij} \), which functions as a gating mechanism. This mask ensures that only easy-to-moderate samples contribute to the loss during the initial training stages, facilitating a more stable and progressive learning trajectory. Our new SPCL loss function is formally defined as:

\begin{equation}
    \mathcal{L}_{SPCL} = \frac{1}{\sum_{i=1}^{|\mathcal{D}|}\sum_{j=1}^{N_i}v_{ij}} \sum_{i=1}^{|\mathcal{D}|}\sum_{j=1}^{N_i}v_{ij}l_{ij}
    \label{eq:spcl-loss}
\end{equation}
where $\mathcal{L}_{SPCL}$ is the new loss, $\sum_{i=1}^{|\mathcal{D}|}\sum_{j=1}^{N_i}v_{ij}$ is the number of easy samples.

Consequently, the uni-modal networks’ parameters are updated based on this filtered loss, ensuring that the model first learns from well-aligned, lower-difficulty samples before gradually incorporating harder samples as training progresses. 

\begin{equation}
    \theta^{m^{(t+1)}} \leftarrow \theta^{m^{(t)}} - \eta\frac{\partial \mathcal{L}_{SPCL}}{\partial \theta^{m^{(t)}}}, \quad m\in\{a, t, v\}
    \label{eq:spcl-upd}
\end{equation}

Overall, the whole training process is described in Algorithm \ref{algo:spcl-pseudo}. 

\begin{algorithm}[t]
    \caption{Pseudo-code for Training a Multimodal ERC Framework with SPCL Integration.}
    \label{algo:spcl-pseudo}
    \begin{algorithmic}
        \State \textbf{Input} $\mathcal{D}=\{U_1, U_2,\dots, U_{|\mathcal{D}|}\}$
        \State Initialize $\lambda\leftarrow\varepsilon$, $\alpha$
        \For{epoch $t$}
            \For{mini-batch $\mathcal{B}$}
                \State Retrieve $z^m_{ij}$, $z^{joint}_{ij}$ (Equation~\ref{eq:uni-pred}), (Equation~\ref{eq:cross-pred})
                \State Calculate utterance loss $l_{ij}$ (Equation~\ref{eq:utt-loss})
                \State Calculate uni-modal score $s^m_i$ (Equation~\ref{eq:uni-score})
                \State Calculate conversation-level score $s_i$ 
                (Equation~\ref{eq:conv-score})
                 \State Calculate difficulty $\rho_{ij}$ (Equation \ref{eq:diffc})
                \State Retrieve mask $v_{ij}$ (Equation~\ref{eq:mask})
                \State Calculate SPCL loss $\mathcal{L}_{SPCL}$ (Equation~\ref{eq:spcl-loss})
                \State Update model parameters with $\mathcal{L}_{SPCL}$ (Equation~\ref{eq:spcl-upd})
            \EndFor
            \State Update threshold $\lambda$ (Equation~\ref{eq:lambda-upd})
        \EndFor
    \end{algorithmic}
    
\end{algorithm}
\section{Experimental Setup}
\label{sec:exp-setting}
\subsection{Datasets}
We conduct experiments on two benchmark datasets for ERC task that support multi-modal, namely: IEMOCAP~\cite{busso2008iemocap}, MELD~\cite{poria2019meld}. Statistics of the two datasets are summarized in Table \ref{tab:data-stats}.

\textbf{IEMOCAP}~\cite{busso2008iemocap}: A dataset of 12-hour video recordings involving 10 actors. This dataset includes 151 dialogues of binary speakers, split into a total of 7,433 utterances. Each utterance is annotated with one of the 6 emotion labels: happy, sad, neutral, angry, excited, or frustrated.

\textbf{MELD}~\cite{poria2019meld}: A dataset deprived from the TV series ``Friends''. This dataset provides 1,433 multi-party dialogues, segmented into 13,709 utterances. These utterances are classified into: happy, sad, angry, scared, disgusted, and surprised, with sentiment intensity ranging from -3 to 3. 

\begin{table}[!ht]
\centering
\caption{Statistics for IEMOCAP and MELD}
\renewcommand{\arraystretch}{1.2} 
\setlength{\tabcolsep}{4.5pt} 
\small
\begin{tabular}{c|ccc|ccc}
\hline
\multirow{2}{*}{Dataset} & \multicolumn{3}{c|}{Dialogues}    & \multicolumn{3}{c}{Utterances}       \\ 
                         & train       & valid     & test    & train       & valid       & test     \\ \hline
IEMOCAP                  & \multicolumn{2}{c}{120} & 31      & \multicolumn{2}{c}{5,810} & 1,623    \\
MELD                     & 1,039       & 114       & 280     & 9,989       & 1,109       & 2,610    \\ \hline
\end{tabular}
\label{tab:data-stats}
\end{table} 
\subsection{Baselines and Evaluation Metrics}
To evaluate the robustness and stability of our proposed method, we incorporate it into 4 existing models for ERC task, namely: DialogueGCN~\cite{ghosal2019dialoguegcn}, BiDDIN~\cite{zhang2020modeling}, MMGCN~\cite{hu2021mmgcn}, MM-DFN~\cite{hu2022mm}. In particular, these models are used as our emotion prediction network $\phi_m(\cdot)$ in Equation~\ref{eq:uni-pred}. 
\begin{itemize}
    \item DialogueGCN~\cite{ghosal2019dialoguegcn}: Models intra- and inter-speaker dependencies using a bidirectional GRU for sequential encoding and a speaker-level graph encoder. Nodes exchange contextual information via similarity-based attention. Originally text-only, we extended it to multimodal by incorporating visual and audio data with late fusion.
    \item BiDDIN~\cite{zhang2020modeling}: Captures intra- and inter-modal dependencies using a bidirectional GRU for modality-specific encoding and a graph-based encoder for cross-modal interactions. Nodes refine representations via message passing, with edge weights set by similarity-based attention. Emotion classification is performed as a node classification task in a multimodal graph.
    \item MMGCN~\cite{hu2021mmgcn}: A multimodal, speaker-aware model using a graph-based fusion framework. A deep GCN refines node embeddings, integrating intra-speaker, inter-speaker, and cross-modal relationships. Through message passing, long-distance contextual information is aggregated for emotion classification as a node classification task.
    \item MM-DFN~\cite{hu2022mm}:  Utilizes a graph-based fusion mechanism for intra- and inter-modal dependencies. A modality-specific encoder processes features separately, while a dynamic fusion module filters redundancy and preserves complementary signals. Emotion classification is framed as a node classification task in a multimodal graph.
\end{itemize}
Since our method requires directly computing uni-modal scores from uni-modal logits, all baselines follow a late-fusion structure. To provide a clearer comparison, we also evaluate our approach against existing frameworks and modules designed to address imbalanced multimodal learning, including RNA loss~\cite{planamente2022domain}, OGM-GE~\cite{peng2022balanced}, and FAGM~\cite{wang2023unlocking}.
Specifically, RNA loss introduces constraints in the loss function to align feature norms across modalities, ensuring more balanced representations. OGM-GE mitigates modality imbalance by dynamically adjusting gradient updates based on the discrepancy ratio. FAGM is a plug-in method that rebalances gradients at the parameter level by proportionally adjusting them based on modality dominance, preventing any single modality from overwhelming the learning process.
While originally developed for bimodal settings, these methods are extended to trimodal models in our experiments. All modules and frameworks are integrated into the baselines and evaluated under the same training environment for a fair comparison.

In order to assess the performance of our model, we employ two key evaluation metrics: the Weighted F1-score (w-F1) and Accuracy (Acc.). These metrics provide insights into the effectiveness and overall correctness of the predictions made by our classifier.

\subsection{Multimodal Raw Feature Extraction}
\label{app:imd}
The multimodal feature extraction process involves extracting features from the acoustic, lexical, and visual modalities for each utterance. For both IEMOCAP and MELD datasets, audio features are obtained using the OpenSmile Toolkit~\cite{eyben2010opensmile}; visual features are extracted using OpenFace~\cite{baltrusaitis2018openface}; textual features are derived using sBERT~\cite{reimers-gurevych-2019-sentence}.
\begin{table}[!ht]
    \centering
    \caption{Hyper-parameters settings}
    \label{tab:hyper-params}
    \renewcommand{\arraystretch}{1.2} 
\setlength{\tabcolsep}{4.5pt} 
\small
    \begin{tabular}{c|cc}
        \hline
        \textbf{Parameter/Module} & \textbf{IEMOCAP} & \textbf{MELD} \\
        \hline
        Text Feature Extraction    & \multicolumn{2}{c}{sBERT\cite{reimers-gurevych-2019-sentence}} \\
        Audio Feature Extraction   & \multicolumn{2}{c}{OpenSmile Toolkit\cite{eyben2010opensmile}} \\
        Visual Feature Extraction  & \multicolumn{2}{c}{OpenFace Toolkit\cite{baltrusaitis2018openface}} \\
        \hline
        Text embedding dim. $d_t$  & 768  & 768   \\
        Audio embedding dim. $d_a$ & 512  & 300   \\
        Visual embedding dim. $d_v$ & 1024 & 342   \\
        \hline
        $\varepsilon$  &              \multicolumn{2}{c}{[0.6, 1.2]}  \\
        
        $\alpha$  &              \multicolumn{2}{c}{[1.05, 1.4]}  \\
        \hline
        Learning-rate               & 
                   \multicolumn{2}{c}{[0.0001, 0.0003]} \\
        Batch size                  & 16   & 32    \\
        Epoch                        & 50   & 50 \\
        \hline
    \end{tabular}
\end{table} 
\subsection{Reproducibility}
\mName{} is implemented using Pytorch \footnote{https://pytorch.org/}, and run experiments on Google Colab and Kaggle. We choose Adam as the optimizer. The batch size is 16 and 32 for IEMOCAP and MELD dataset, respectively. Since each combination of baseline and dataset have different converging rates, the hyper-parameters are tested on various settings. Particularly, learning-rate is selected within the range of $[0.0001, 0.0003]$; hyper-parameter $\varepsilon$, i.e. initial value of threshold $\lambda$, is picked from range of $[0.6,1.2]$; raging hyper-parameter $\alpha$ is selected from range of $[1.05, 1.4]$.
The hyper-parameter settings used in the experiments are presented in Table \ref{tab:hyper-params}.

\section{Result and Discussion}
\label{sec:results}
We qualitatively analyze our proposed Self-paced Curriculum Learning-based Approach (\mName{}) and the baselines on the
IEMOCAP and MELD datasets. We also conducted extensive experiments to prove the utility of each individual components of the Difficulty Measurer in the ablation study section.
\subsection{Comparision with Baselines}
\subsubsection{Analysis of Experimental Results on IEMOCAP}  

Table~\ref{tab:iemocap-result} presents a comparative performance analysis of our proposed \mName{} module against multiple baselines on the IEMOCAP dataset. The results demonstrate that integrating \mName{} consistently improves weighted F1-score (w-F1) and accuracy (Acc) across all modality combinations (TAV, TA, TV, AV), outperforming existing methods. The performance gap with other imbalance-mitigation methods (\(\Delta\)) and the improvement over the original baseline model without any balancing strategy (\(\Delta_{Base}\)) highlight the effectiveness of \mName{}. 
\begin{table}[!ht]
\centering
\caption{Performance comparison of baseline models with our SPCL module and other plug-in methods on IEMOCAP. \textbf{Bold} and \uline{underlined} denote the best and second-best results, respectively. $\Delta$ indicates the performance gap to the previous SOTA, while $\Delta_{\text{Base}}$ measures the improvement of SPCL over the original baseline. Values marked with $^{\dagger}$ denote statistically significant improvements ($p < 0.05$) based on paired t-tests.}

\label{tab:iemocap-result}
\renewcommand{\arraystretch}{1.45}
\setlength{\tabcolsep}{1.5pt}
\footnotesize
\begin{tabular}{l|cc|cc|cc|cc}
\hline
\multirow{2}{*}{Model} & \multicolumn{2}{c|}{TAV} & \multicolumn{2}{c|}{TA} & \multicolumn{2}{c|}{TV} & \multicolumn{2}{c}{AV} \\
 & w-F1 & Acc & w-F1 & Acc & w-F1 & Acc & w-F1 & Acc \\ \hline
\multicolumn{9}{c}{\textit{DialogueGCN}~\cite{ghosal2019dialoguegcn}} \\ \hline
Baseline & 60.43 & 60.54 & 61.61 & 61.72 & 59.19 & 59.48 & 47.89 & 48.49 \\
+ RNA loss & 58.43 & 58.47 & 57.42 & 57.73 & 56.23 & 56.62 & 47.40 & 49.29 \\
+ OGM-GE & 57.16 & 57.24 & 59.30 & 59.52 & 55.88 & 56.13 & 43.71 & 44.98 \\
+ OPM & 58.89 & 59.72 & 57.02 & 57.55 & 60.48 & 60.54 & \uline{49.80} & \uline{51.76} \\
+ FAGM & \uline{62.76} & \uline{63.22} & \uline{64.36} & \uline{64.39} & \uline{61.25} & \uline{62.23} & 49.20 & 49.85 \\
+ \textbf{SPCL} & \textbf{66.99}$^{\dagger}$\std{1.03} & \textbf{67.03}$^{\dagger}$ & \textbf{65.32}$^{\dagger}$\std{0.99} & \textbf{65.46}$^{\dagger}$ & \textbf{64.47}$^{\dagger}$\std{0.21} & \textbf{64.46}$^{\dagger}$ & \textbf{57.89}$^{\dagger}$\std{1.00} & \textbf{58.59}$^{\dagger}$ \\
\rowcolor{lightgray} \qquad $\Delta$ & \textit{4.23} & \textit{3.81} & \textit{0.96} & \textit{1.07} & \textit{3.22} & \textit{2.23} & \textit{8.09} & \textit{6.83} \\
\rowcolor{lightgray} \qquad $\Delta_{\text{Base}}$ & \textit{6.56} & \textit{6.49} & \textit{3.71} & \textit{3.74} & \textit{5.28} & \textit{4.98} & \textit{10.00} & \textit{10.10} \\ \hline

\multicolumn{9}{c}{\textit{BiDDIN}~\cite{zhang2020modeling}} \\ \hline
Baseline & 58.29 & 58.20 & 58.73 & \uline{58.67} & 58.57 & 57.93 & 45.35 & 46.03 \\
+ RNA loss & 58.63 & 58.55 & 58.02 & 57.92 & 57.29 & 57.24 & 42.54 & 44.82 \\
+ OGM-GE & 58.06 & 57.98 & 57.71 & 57.73 & 57.58 & 57.55 & 39.84 & 40.42 \\
+ OPM & 56.27 & 56.62 & 57.82 & 57.60 & 52.59 & 52.60 & 37.72 & 40.48 \\
+ FAGM & \uline{58.81} & \uline{58.84} & \uline{58.88} & 58.16 & \uline{59.04} & \uline{58.96} & \textbf{46.36} & \uline{46.77} \\
+ \textbf{SPCL} & \textbf{59.90}$^{\dagger}$\std{0.13} & \textbf{60.73}$^{\dagger}$ & \textbf{60.24}$^{\dagger}$\std{1.11} & \textbf{60.43}$^{\dagger}$ & \textbf{61.10}$^{\dagger}$\std{0.82} & \textbf{61.91}$^{\dagger}$ & \uline{46.34}\std{0.43} & \textbf{49.11}$^{\dagger}$ \\
\rowcolor{lightgray} \qquad $\Delta$ & \textit{1.09} & \textit{1.89} & \textit{1.36} & \textit{1.76} & \textit{2.06} & \textit{2.95} & \textit{-0.02} & \textit{2.34} \\
\rowcolor{lightgray} \qquad $\Delta_{\text{Base}}$ & \textit{1.61} & \textit{2.53} & \textit{1.51} & \textit{1.76} & \textit{2.53} & \textit{3.98} & \textit{0.99} & \textit{3.08} \\ \hline

\multicolumn{9}{c}{\textit{MMGCN}~\cite{hu2021mmgcn}} \\ \hline
Baseline & 62.67 & 62.67 & 62.66 & 62.72 & 58.99 & 59.14 & 47.22 & 49.23 \\
+ RNA loss & 63.13 & 63.28 & 59.25 & 59.27 & 56.30 & 56.50 & 50.35 & 51.20 \\
+ OGM-GE & 62.42 & 62.69 & 62.33 & 62.42 & 58.83 & 59.03 & 51.90 & 53.54 \\
+ OPM & \uline{64.60} & 64.10 & 62.30 & 62.70 & 59.70 & 59.60 & 50.60 & 52.00 \\
+ FAGM & 64.53 & \uline{64.51} & \uline{63.25} & \uline{63.40} & \uline{61.02} & \uline{61.06} & \textbf{54.14} & \textbf{54.90} \\
+ \textbf{SPCL} & \textbf{67.66}$^{\dagger}$\std{0.57} & \textbf{67.71}$^{\dagger}$ & \textbf{66.75}$^{\dagger}$\std{0.42} & \textbf{66.51}$^{\dagger}$ & \textbf{65.00}$^{\dagger}$\std{1.05} & \textbf{65.09}$^{\dagger}$ & \uline{53.70}\std{0.71} & \uline{54.04}\std{0.94} \\
\rowcolor{lightgray} \qquad $\Delta$ & \textit{3.06} & \textit{3.20} & \textit{3.50} & \textit{3.11} & \textit{3.98} & \textit{4.03} & \textit{-0.44} & \textit{-0.86} \\
\rowcolor{lightgray} \qquad $\Delta_{\text{Base}}$ & \textit{5.00} & \textit{5.04} & \textit{4.09} & \textit{3.79} & \textit{6.01} & \textit{5.95} & \textit{6.48} & \textit{4.81} \\ \hline

\multicolumn{9}{c}{\textit{MM-DFN}~\cite{hu2022mm}} \\ \hline
Baseline & 61.54 & 61.72 & 61.98 & 62.12 & 59.78 & 59.93 & 48.42 & 49.11 \\
+ RNA loss & 60.23 & 60.49 & 60.18 & 60.41 & 57.74 & 57.92 & 45.63 & 46.32 \\
+ OGM-GE & 59.92 & 60.13 & 60.57 & 60.69 & 58.33 & 58.49 & 44.98 & 45.51 \\
+ OPM & 63.30 & 62.91 & \uline{64.43} & \uline{64.45} & \uline{64.06} & \uline{63.89} & \uline{53.55} & \uline{53.79} \\
+ FAGM & \uline{63.45} & \uline{63.72} & 63.83 & 63.94 & 61.58 & 61.72 & 50.35 & 51.02 \\
+ \textbf{SPCL} & \textbf{67.16}$^{\dagger}$\std{0.67} & \textbf{67.08}$^{\dagger}$ & \textbf{66.03}$^{\dagger}$\std{0.86} & \textbf{66.09}$^{\dagger}$ & \textbf{64.31}$^{\dagger}$\std{0.66} & \textbf{64.70}$^{\dagger}$ & \textbf{53.38}$^{\dagger}$\std{0.67} & \textbf{53.47}$^{\dagger}$ \\
\rowcolor{lightgray} \qquad $\Delta$ & \textit{3.71} & \textit{3.36} & \textit{1.60} & \textit{1.64} & \textit{0.25} & \textit{0.81} & \textit{-0.17} & \textit{-0.32} \\
\rowcolor{lightgray} \qquad $\Delta_{\text{Base}}$ & \textit{5.62} & \textit{5.36} & \textit{4.05} & \textit{3.97} & \textit{4.53} & \textit{4.77} & \textit{4.96} & \textit{4.36} \\ \hline
\end{tabular}
\end{table}

\textbf{Overall Performance Improvements:} Across all baseline models, our method achieves state-of-the-art performance, yielding the highest accuracy and weighted F1 scores, with statistically significant improvements over the strongest existing approach, such as FAGM. Notably, our approach demonstrates substantial gains in TAV and AV settings, where modality imbalance poses a significant challenge. Compared to the baseline without any balancing strategy, our method consistently delivers marked performance enhancements. In the DialogueGCN (TAV) setting, the baseline achieves 60.43\% w-F1 and 60.54\% accuracy, whereas our method significantly improves these to 66.99\% w-F1 (+6.56\%) and 67.03\% accuracy (+6.49\%). Similarly, in MM-DFN (TAV), our method surpasses the baseline by 5.62\% in w-F1 and 5.66\% in accuracy. The improvements are also consistent in the AV setting, where our method achieves 58.30\% w-F1 and 58.47\% accuracy for DialogueGCN, representing gains of +10.41\% w-F1 and +9.98\% accuracy over the baseline.

While FAGM achieves competitive performance in some cases, other methods such as RNA loss and OGM-GE frequently result in performance degradation. For instance, in DialogueGCN (TAV), RNA loss reduces w-F1 from 60.43\% to 58.43\% and accuracy from 60.54\% to 58.47\%. OGM-GE further degrades performance to 57.16\% w-F1 and 57.24\% accuracy. This indicates the limitations of static regularization approaches in handling modality imbalance. OPM yields only modest and inconsistent improvements over the baseline. For instance, in MM-DFN (AV), it raises w-F1 from 53.30\% to 54.02\%, yet remains 4.28\% below our method's 58.30\%. This indicates that fixed reweighting strategies like OPM are insufficient for capturing dynamic modality contributions under imbalance.

These results suggest that the success of our method stems from its ability to dynamically adapt to the evolving learning difficulty of samples and the shifting contributions of different modalities. Unlike static or manually designed weighting schemes, our SPCL framework leverages real-time feedback from both utterance-level performance and conversation-level modality discrepancies. This dual-level perspective enables the model to prioritize informative yet underrepresented modalities and to avoid overfitting to dominant signals. As a result, the training process becomes more balanced and effective, leading to superior generalization performance across various MERC settings.

\textbf{Impact on Modality Combinations:}  
Among different modality combinations, the TAV setting exhibits the most substantial improvements with SPCL, effectively addressing modality imbalance. Across models, SPCL outperforms FAGM, achieving w-F1 gains ranging from 1.17\% to 3.70\%, demonstrating the benefits of adaptive sample selection in enhancing multimodal alignment. The TA and TV settings also experience consistent improvements, particularly in MMGCN when integrating SPCL compared to integrating FAGM, where accuracy increases from 63.26\% to 66.15\% (+2.89\%), and in MM-DFN, where it improves from 63.94\% to 66.80\% (+2.86\%). This suggests that SPCL effectively strengthens the interaction between textual and non-textual modalities.

The AV setting, which poses the greatest challenge due to the absence of textual features, exhibits the most pronounced improvements. In DialogueGCN, SPCL surpasses FAGM, improving w-F1 from 49.20\% to 57.98\% and accuracy from 49.85\% to 58.49\%, achieving gains of 8.78\% and 8.64\%, respectively. Similarly, in MM-DFN, SPCL enhances w-F1 from 50.04\% to 58.30\% and accuracy from 50.91\% to 58.47\%, with improvements of 8.26\% and 7.56\%. These findings highlight the robustness of SPCL in optimizing non-textual modality fusion, making it particularly effective in overcoming modality imbalance.

\subsubsection{Analysis of Experimental Results on MELD} 
Table~\ref{tab:meld-result} presents a comparative performance analysis of our proposed \mName{} module against multiple baselines on the MELD dataset. Similar to IEMOCAP, integrating \mName{} consistently improves weighted F1-score (w-F1) and accuracy (Acc) across all modality combinations (TAV, TA, TV, AV), surpassing existing approaches.

\begin{table}[!ht]
\centering
\caption{Performance comparison of baseline models with our SPCL module and other plug-in methods on MELD. \textbf{Bold} and \uline{underlined} denote the best and second-best results, respectively. $\Delta$ indicates the performance gap to the previous SOTA, while $\Delta_{\text{Base}}$ measures the improvement of SPCL over the original baseline. Values marked with $^{\dagger}$ denote statistically significant improvements ($p < 0.05$) based on paired t-tests.}

\label{tab:meld-result}
\renewcommand{\arraystretch}{1.45}
\setlength{\tabcolsep}{1.5pt}
\footnotesize
\begin{tabular}{l|cc|cc|cc|cc}
\hline
\multirow{2}{*}{Model} & \multicolumn{2}{c|}{TAV} & \multicolumn{2}{c|}{TA} & \multicolumn{2}{c|}{TV} & \multicolumn{2}{c}{AV} \\
 & w-F1 & Acc & w-F1 & Acc & w-F1 & Acc & w-F1 & Acc \\ \hline
\multicolumn{9}{c}{\textit{DialogueGCN}~\cite{ghosal2019dialoguegcn}} \\ \hline
Baseline & 53.11 & 55.08 & 51.99 & 54.22 & 54.22 & 56.07 & \uline{43.54} & 44.54 \\
+ RNA loss & 56.65 & 58.47 & 54.21 & 58.35 & 53.78 & \uline{58.12} & \textbf{43.64} & \textbf{47.32} \\
+ OGM-GE & \uline{57.73} & 57.36 & \textbf{56.38} & \textbf{58.81} & \uline{56.15} & 57.78 & 42.05 & 46.51 \\
+ OPM & 54.47 & 57.12 & 53.26 & 56.17 & 53.21 & 57.66 & 40.52 & 43.64 \\
+ FAGM & 54.61 & \uline{58.96} & \uline{54.80} & 57.28 & 55.26 & 57.10 & 40.02 & 44.44 \\
+ \textbf{SPCL} & \textbf{57.87}\std{1.49} & \textbf{60.77}$^{\dagger}$ & \uline{58.04}\std{0.56} & \uline{60.84} & \textbf{56.18}\std{1.38} & \textbf{58.61}$^{\dagger}$ & 42.28\std{0.79} & \uline{46.64} \\
\rowcolor{lightgray} \qquad$\Delta$ & \textit{0.14} & \textit{1.81} & \textit{1.66} & \textit{2.03} & \textit{0.03} & \textit{0.49} & \textit{-1.36} & \textit{-0.68} \\
\rowcolor{lightgray} \qquad$\Delta_{\text{Base}}$ & \textit{4.76} & \textit{5.69} & \textit{6.05} & \textit{6.62} & \textit{1.96} & \textit{2.54} & \textit{-1.26} & \textit{2.10} \\ \hline

\multicolumn{9}{c}{\textit{BiDDIN}~\cite{zhang2020modeling}} \\ \hline
Baseline & 56.41 & 58.54 & 56.23 & 57.85 & 56.46 & 58.06 & \uline{43.07} & 47.35 \\
+ RNA loss & 52.18 & 49.16 & 53.21 & 50.31 & 52.59 & 49.43 & 41.05 & 44.60 \\
+ OGM-GE & 55.27 & 53.41 & 51.96 & 47.74 & 52.18 & 48.58 & 43.03 & 46.97 \\
+ OPM & 53.87 & 57.62 & 54.73 & \uline{58.58} & 56.25 & \uline{59.77} & 40.69 & 47.39 \\
+ FAGM & \uline{57.47} & \uline{59.18} & \uline{56.56} & 58.05 & \uline{56.93} & 58.10 & \textbf{44.39} & \textbf{48.62} \\
+ \textbf{SPCL} & \textbf{57.60}$^{\dagger}$\std{0.25} & \textbf{60.86}$^{\dagger}$ & \textbf{58.08}$^{\dagger}$\std{0.30} & \textbf{61.22}$^{\dagger}$ & \textbf{58.10}$^{\dagger}$\std{0.43} & \textbf{61.00}$^{\dagger}$ & 42.30\std{0.23} & \uline{48.15} \\
\rowcolor{lightgray} \qquad$\Delta$ & \textit{0.13} & \textit{1.68} & \textit{1.52} & \textit{2.64} & \textit{1.17} & \textit{1.23} & \textit{-2.09} & \textit{-0.47} \\
\rowcolor{lightgray} \qquad$\Delta_{\text{Base}}$ & \textit{1.19} & \textit{2.32} & \textit{1.85} & \textit{3.37} & \textit{1.64} & \textit{2.94} & \textit{-0.77} & \textit{1.12} \\ \hline

\multicolumn{9}{c}{\textit{MMGCN}~\cite{hu2021mmgcn}} \\ \hline
Baseline & 57.71 & 59.95 & 57.29 & 59.79 & 56.73 & 59.31 & 42.38 & \textbf{49.12} \\
+ RNA loss & 56.94 & 58.62 & 56.00 & 57.59 & 55.48 & 57.70 & 41.84 & 46.91 \\
+ OGM-GE & 57.59 & 59.92 & 56.80 & 59.77 & 56.20 & 59.08 & 42.20 & 48.81 \\
+ OPM & 55.78 & 57.24 & 56.27 & 59.77 & 55.29 & 59.23 & 42.72 & 47.20 \\
+ FAGM & \uline{58.48} & \uline{61.15} & \uline{57.59} & \uline{60.69} & \uline{57.14} & \uline{59.46} & \uline{43.49} & 48.43 \\
+ \textbf{SPCL} & \textbf{59.11}$^{\dagger}$\std{0.48} & \textbf{61.32}$^{\dagger}$ & \textbf{58.93}$^{\dagger}$\std{0.29} & \textbf{61.65}$^{\dagger}$ & \textbf{58.14}$^{\dagger}$\std{1.17} & \textbf{60.64}$^{\dagger}$ & \textbf{43.79}$^{\dagger}$\std{0.31} & \uline{49.10} \\
\rowcolor{lightgray} \qquad$\Delta$ & \textit{0.63} & \textit{0.17} & \textit{1.34} & \textit{0.96} & \textit{1.00} & \textit{1.18} & \textit{0.30} & \textit{-0.02} \\
\rowcolor{lightgray} \qquad$\Delta_{\text{Base}}$ & \textit{1.40} & \textit{1.37} & \textit{1.64} & \textit{1.86} & \textit{1.41} & \textit{1.33} & \textit{1.41} & \textit{-0.02} \\ \hline

\multicolumn{9}{c}{\textit{MM-DFN}~\cite{hu2022mm}} \\ \hline
Baseline & 57.52 & 59.90 & 57.11 & 59.47 & 57.46 & 59.68 & 40.04 & 43.91 \\
+ RNA loss & 56.02 & 58.20 & 54.13 & 55.59 & 54.13 & 55.59 & 36.39 & 47.54 \\
+ OGM-GE & 56.53 & 58.39 & 55.86 & 59.08 & 56.25 & 58.24 & 40.60 & 48.43 \\
+ OPM & \uline{58.75} & \uline{61.42} & \uline{57.67} & \uline{61.38} & \uline{58.28} & \uline{61.49} & \textbf{42.51} & 47.16 \\
+ FAGM & 57.55 & 60.80 & 57.10 & 60.00 & 57.73 & 60.65 & 42.05 & \textbf{48.66} \\
+ \textbf{SPCL} & \textbf{59.17}$^{\dagger}$\std{0.30} & \textbf{61.91}$^{\dagger}$ & \textbf{59.11}$^{\dagger}$\std{0.32} & \textbf{62.31}$^{\dagger}$ & \textbf{58.91}$^{\dagger}$\std{0.17} & \textbf{61.94}$^{\dagger}$ & \uline{43.32}\std{0.57} & \uline{48.59}$^{\dagger}$ \\
\rowcolor{lightgray}\qquad $\Delta$ & \textit{0.42} & \textit{0.49} & \textit{1.44} & \textit{0.93} & \textit{0.63} & \textit{0.45} & \textit{0.81} & \textit{-0.07} \\
\rowcolor{lightgray} \qquad$\Delta_{\text{Base}}$ & \textit{1.65} & \textit{2.01} & \textit{2.00} & \textit{2.84} & \textit{1.45} & \textit{2.26} & \textit{3.28} & \textit{4.68} \\ \hline
\end{tabular}
\end{table}

\textbf{Overall Performance Improvements:} Our method consistently achieves the highest performance across all baseline models in the TAV setting, outperforming competitive approaches such as FAGM. For example, in MM-DFN, with SPCL integrated, our method improves the weighted F1-score from 57.55\% (FAGM) to 59.17\% (+1.62\%) and from the baseline’s 57.52\%, yielding a total gain of +1.65\%. Similarly, in MMGCN, SPCL increases the weighted F1-score from 58.48\% (FAGM) to 59.11\% (+0.63\%) and over the baseline’s 57.71\%, achieving a total improvement of +1.40\%. Across all evaluated models in the TAV setting, SPCL achieves an average weighted F1-score improvement of 0.85\% over the second-best method and 2.25\% over the baseline models, demonstrating consistent effectiveness in enhancing multimodal interactions.

While FAGM remains competitive, SPCL demonstrates a more adaptive learning strategy, particularly within transformer-based architectures. For instance, in MM-DFN on MELD, SPCL surpasses the second-best method OPM by 0.42\% in the TAV setting (59.17\% vs. 58.75\%). Consistent with findings on IEMOCAP, static regularization techniques such as RNA loss and OGM-GE often fail to deliver consistent performance improvements. While RNA loss improves performance in certain cases (e.g., 56.65\% weighted F1-score in DialogueGCN’s TAV setting), it does not consistently achieve the best results across different models.

However, in DialogueGCN on MELD, our method does not consistently yield superior performance. In the TAV setting, SPCL achieves a weighted F1-score of 57.87\%, which is only 0.14\% higher than OGM-GE (57.73\%). The limited effectiveness of our curriculum-based training on MELD may be attributed to the dataset’s shorter and more fragmented conversational structure. As SPCL progressively introduces more complex samples, its training schedule might not align optimally with MELD’s data distribution, thereby limiting its potential gains within this specific architecture.

\textbf{Impact of Different Modality Combinations:}
The performance trends across modality combinations on MELD are largely consistent with those observed on IEMOCAP, further validating the effectiveness of our proposed approach. The TAV setting particularly benefits from SPCL, as its adaptive sample selection enhances multimodal balance and improves overall recognition performance. Additionally, the TA and TV settings exhibit notable improvements, demonstrating the capacity of SPCL to mitigate modality imbalance across diverse multimodal configurations.

Similar to IEMOCAP, SPCL consistently outperforms FAGM across models in the TA, TV, and AV settings. In the TA setting, SPCL achieves weighted F1-score improvements ranging from 1.39\% to 1.75\% over FAGM, with the most pronounced gains observed in BiDDIN (+2.08\%) and MM-DFN (+2.41\%) relative to the baseline. In the TV setting, SPCL maintains superior performance, particularly in BiDDIN (+3.54\%) and MMGCN (+2.04\%) over the baseline model. For the AV setting, while the improvements over competing methods are more moderate, SPCL attains the highest weighted F1-score in MM-DFN (42.42\%) and MMGCN (44.34\%), with notable gains in MM-DFN (+2.38\%) compared to the baseline. These findings further reinforce the efficacy of SPCL in addressing modality imbalance and enhancing multimodal emotion recognition in conversations.




\subsection{Discussion and Analysis}
In this section, we provide further analysis and insights into the effectiveness of our proposed SPCL framework.
\subsubsection{Impact of Key Components}
We conduct an ablation study to evaluate the impact of the two key components in our Difficulty Measurer: the utterance-level score $l_{ij}$ and the conversation-level score $s_i$. Specifically, we systematically remove each component from the difficulty formulation of $\rho_{ij}$ in Equation~\ref{eq:diffc} and assess the resulting performance, as summarized in Table~\ref{tab:ablation-module}.

\begin{table}[!ht]
\centering
\caption{Ablation study on IEMOCAP for our proposed \mName{} module. The subscript $\downarrow$ or $\uparrow$ denotes the performance change compared to our SPCL module when a sub-module is ablated.}
\renewcommand{\arraystretch}{1.2}
\setlength{\tabcolsep}{3pt}
\begin{tabular}{l|cc|cc|cc|cc}
\hline
\multirow{2}{*}{\textbf{Method}} & \multicolumn{2}{c|}{TAV} & \multicolumn{2}{c|}{TA} & \multicolumn{2}{c|}{TV} & \multicolumn{2}{c}{AV} \\
 & w-F1 & Acc & w-F1 & Acc & w-F1 & Acc & w-F1 & Acc \\
\hline
\textbf{DialogueGCN}~\cite{ghosal2019dialoguegcn} & 60.43 & 60.54 & 61.61 & 61.72 & 59.19 & 59.48 & 47.89 & 48.49 \\
\textbf{+ SPCL (Ours)} & \textbf{66.99} & \textbf{67.03} & \textbf{65.32} & 65.46 & \textbf{64.47} & \textbf{64.46} & \textbf{57.89} & \textbf{58.59} \\
\quad w/o utt-score & 63.11$_{\downarrow 3.88}$ & 63.24 & 65.31$_{\downarrow 0.01}$ & 65.72 & 63.56$_{\downarrow 0.91}$ & 63.72 & 55.29$_{\downarrow 2.60}$ & 56.13 \\
\quad w/o conv-score & 64.59$_{\downarrow 2.40}$ & 64.94 & 64.87$_{\downarrow 0.45}$ & \textbf{65.66} & 63.49$_{\downarrow 0.98}$ & 63.86 & 55.25$_{\downarrow 2.64}$ & 56.06 \\
\hline
\textbf{BiDDIN}~\cite{zhang2020modeling} & 58.29 & 58.20 & 58.73 & 58.67 & 58.57 & 57.93 & 45.35 & 46.03 \\
\textbf{+ SPCL (Ours)} & \textbf{59.90} & \textbf{60.73} & 59.40 & 60.24 & \textbf{61.10} & \textbf{61.91} & \textbf{46.34} & \textbf{49.11} \\
\quad w/o utt-score & 57.59$_{\downarrow 2.31}$ & 59.18 & \textbf{60.41}$_{\uparrow 1.01}$ & \textbf{60.59} & 60.67$_{\downarrow 0.43}$ & 61.28 & 45.02$_{\downarrow 1.32}$ & 48.31 \\
\quad w/o conv-score & 58.61$_{\downarrow 1.29}$ & 59.22 & 59.14$_{\downarrow 0.26}$ & 60.46 & 59.48$_{\downarrow 1.62}$ & 60.08 & 45.41$_{\downarrow 0.93}$ & 48.50 \\
\hline
\textbf{MMGCN}~\cite{hu2021mmgcn} & 62.67 & 62.67 & 62.66 & 62.72 & 58.99 & 59.14 & 47.22 & 49.23 \\
\textbf{+ SPCL (Ours)} & \textbf{67.66} & \textbf{67.71} & 65.62 & 65.84 & 66.01 & 65.91 & \textbf{53.70} & 54.04 \\
\quad w/o utt-score & 64.60$_{\downarrow 3.06}$ & 65.05 & 63.89$_{\downarrow 1.73}$ & 64.01 & 62.55$_{\downarrow 3.46}$ & 62.75 & 50.31$_{\downarrow 3.39}$ & 52.18 \\
\quad w/o conv-score & 65.78$_{\downarrow 1.88}$ & 65.85 & \textbf{66.02}$_{\uparrow 0.40}$ & \textbf{65.98} & \textbf{66.11}$_{\uparrow 0.10}$ & \textbf{66.07} & 52.00$_{\downarrow 1.70}$ & \textbf{55.32} \\
\hline
\textbf{MM-DFN}~\cite{hu2022mm} & 61.84 & 61.84 & 61.95 & 62.04 & 60.32 & 60.37 & 50.96 & 52.87 \\
\textbf{+ SPCL (Ours)} & \textbf{67.16} & \textbf{67.08} & \textbf{66.09} & \textbf{66.51} & \textbf{65.43} & \textbf{64.91} & \textbf{53.38} & \textbf{57.40} \\
\quad w/o utt-score & 66.46$_{\downarrow 0.70}$ & 66.39 & 65.18$_{\downarrow 0.91}$ & 65.45 & 63.73$_{\downarrow 1.70}$ & 63.94 & 52.11$_{\downarrow 1.27}$ & 52.53 \\
\quad w/o conv-score & 64.49$_{\downarrow 2.67}$ & 64.54 & 65.20$_{\downarrow 0.89}$ & 65.45 & 63.95$_{\downarrow 1.48}$ & 64.29 & 50.93$_{\downarrow 2.45}$ & 53.13 \\
\hline
\end{tabular}
\label{tab:ablation-module}
\end{table}

Overall, across all baseline models and modality settings, removing either component leads to consistent performance degradation, confirming their complementary roles in SPCL. The performance drop is particularly pronounced in the TAV setting, which involves the full modality set and exhibits more complex inter-modal dynamics. For instance, in DialogueGCN (TAV), removing the utterance-level score causes a drop of 3.88\% in weighted F1, while removing the conversation-level score results in a 2.67\% decrease.

The utterance-level score proves to be especially critical, as its removal leads to substantial and consistent performance drops across multiple models and settings (e.g., $-3.06\%$ in MMGCN (TAV), $-1.70\%$ in MMDFN (TV)). This highlights its importance in capturing fine-grained, modality-specific discrepancies at the local level, thus guiding SPCL in effective pacing and intra-utterance balancing.

Conversely, the conversation-level score contributes to modeling broader, global-level patterns such as turn-wise modality shifts or long-range emotional dependencies. While its removal leads to smaller declines compared to the utterance-level score, it still yields meaningful gains when present (e.g., $+2.64\%$ in DialogueGCN (AV), $+1.62\%$ in BiDDIN (TV)).

In a few cases, using only one of the two difficulty scores still surpasses the baseline performance (e.g., BiDDIN (TV) without conv-score achieves $+0.91\%$ over the baseline), underscoring the independent utility of each score. However, the full SPCL module consistently achieves the best results across all cases, reaffirming that both levels of difficulty modeling are necessary for addressing the diverse imbalance patterns in MERC.




\subsubsection{Curricula Expanding Rate and Hyper-parameters Tuning}
\label{sec:tuning}
\begin{figure}[!t]
    \centering
\includegraphics[width=0.67\textwidth]{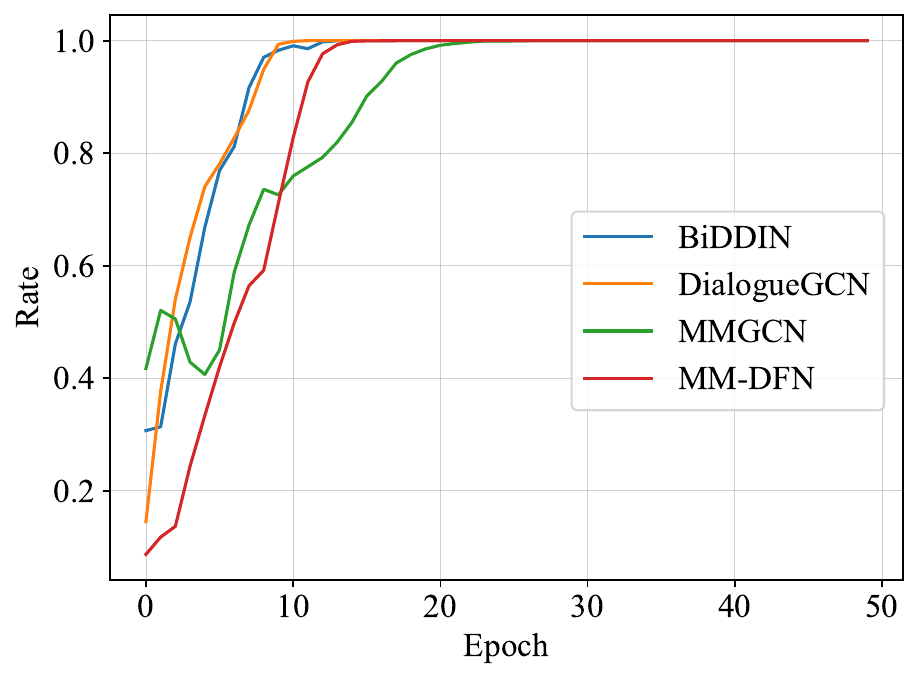}
    \caption{The curricula expanding rate of the four baselines integrated on IEMOCAP.}
    \label{fig:expanding-rate}
\end{figure}

We define the curriculum expanding rate as the ratio of easy samples to the total samples at each training epoch. This rate ranges between 0 and 1, where a value of 1 indicates training on the entire dataset. However, it is not guaranteed to increase consistently unless carefully tuned. The expanding rates of various baselines, when integrated with our module, are illustrated in Figure \ref{fig:expanding-rate}. 

This rate is directly influenced by the tuning of $\varepsilon$ and $\alpha$, which can be explained through the updating of $\lambda$ in Equation. \ref{eq:lambda-upd}, and varies depending on the baseline architecture, as different models exhibit unique sensitivity to data distribution.

Our study on the curriculum expanding rate reveals that the best performance is achieved when the rate maintains a consistently increasing trend, as exemplified by DialogueGCN. This suggests that a gradual yet steady introduction of complex samples enhances the learning progression of the model. Furthermore, this expanding rate highlights the critical role of early training phases in shaping overall model performance.

\begin{figure}[!ht]
    \centering

    \begin{minipage}[t]{0.48\textwidth}
        \centering
        \includegraphics[width=\textwidth]{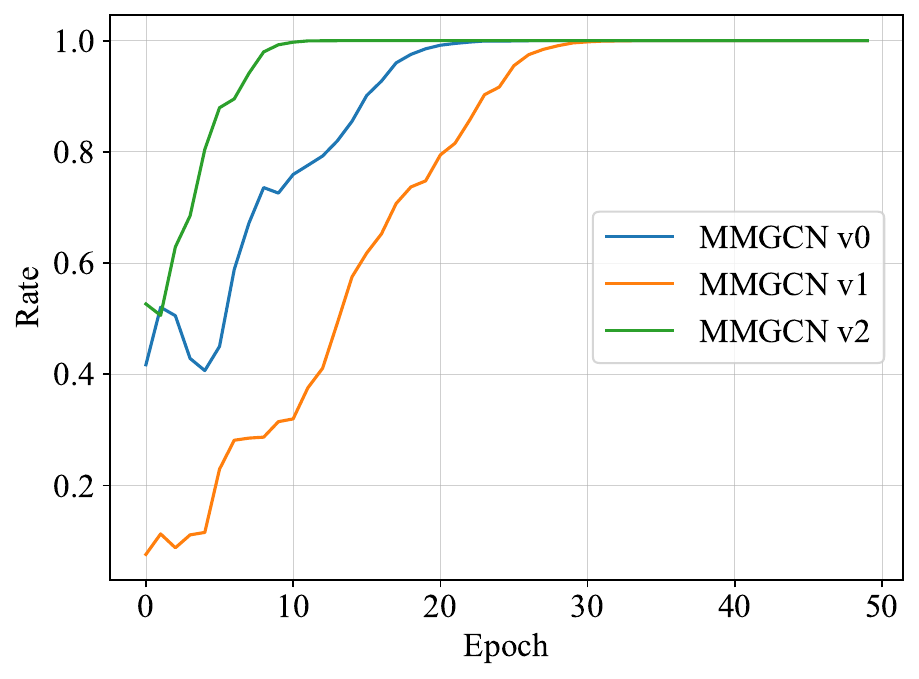}
        \par\small (a) MMGCN
    \end{minipage}
    \hfill
    \begin{minipage}[t]{0.48\textwidth}
        \centering
        \includegraphics[width=\textwidth]{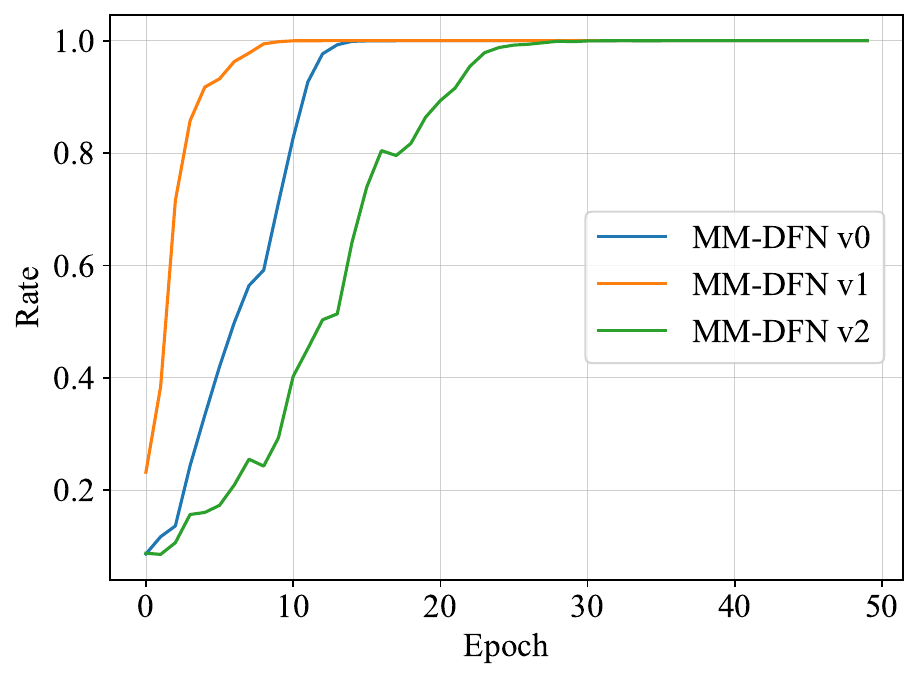}
        \par\small (b) MM-DFN
    \end{minipage}

    \caption{Curricula expanding rate of MMGCN and MM-DFN under \textit{SPCL hyper-parameters setting} specified in Table~\ref{tab:tuning_results}.}
    \label{fig:tuning_results}
\end{figure}

\begin{table}[!ht]
\centering
\caption{Performance of MMGCN and MM-DFN on IEMOCAP under different hyper-parameter settings for our \mName{} module, with v0 representing the best configuration.}
\renewcommand{\arraystretch}{1.2} 
\setlength{\tabcolsep}{4pt} 
\begin{tabular}{lc|c|c|c|c}
\hline
\textbf{Model} & \textbf{Version} & $\boldsymbol{\varepsilon}$ & $\boldsymbol{\alpha}$ & \textbf{w-F1 (\%)} & \textbf{Acc (\%)} \\ 
\midrule
\multirow{3}{*}{\textbf{MMGCN}} 
 & v0 & 0.8 & 1.1 & \textbf{67.84} & \textbf{67.84} \\ 
 & v1 & 0.4 & 1.1 & 67.19 & 64.02 \\  
 & v2 & 0.8 & 1.2 & 65.84 & 65.56 \\  
\hline
\multirow{3}{*}{\textbf{MM-DFN}} 
 & v0 & 0.4 & 1.2 & \textbf{67.92} & \textbf{68.21} \\  
 & v1 & 0.8 & 1.2 & 67.45 & 67.80 \\  
 & v2 & 0.6 & 1.1 & 66.83 & 66.51 \\  
\hline
\end{tabular}
\label{tab:tuning_results}
\end{table}
Since hyperparameter tuning is crucial, we further investigate this by conducting an ablation study on MMGCN and MMDFN using the IEMOCAP dataset. We experiment with different hyperparameter settings and analyze their impact on the curriculum expanding rate, as illustrated in the corresponding Table \ref{tab:tuning_results} and Figure \ref{fig:tuning_results}. Our findings indicate that $\varepsilon$ and $\alpha$ are proportional to the expanding rate, meaning that the rate can be sped up by increasing these values or slowed down by decreasing them.

From our experiments, we observe that each model is optimized for a specific expanding rate. The v0 setting yields the best performance, whereas both speeding up and slowing down (v1+v2) the expanding rate result in performance degradation. 
Our intuitive explanation for this phenomenon is that if the expanding rate is too fast, weak modalities with slower learning rates will fail to fully exploit easy samples, leading to an unreliable starting point and degrading the training process later on. Conversely, if the expanding rate is too slow, the model tends to overfit on easy samples and struggles to learn from hard examples due to mismatched data distribution, ultimately resulting in poor generalization.

\subsubsection{Analysis of Regularization Strategy}
We conducted a comprehensive comparison between our proposed hard regularizer and two alternative soft regularization strategies, namely the Linear and Logistic regularizers. The implementations of these soft regularizers follow the closed-form formulations described in~\cite{wang2021survey}.
\begin{table}[!ht]
\centering
\caption{Performance comparison of four backbone models on IEMOCAP and MELD datasets using different types of regularizers for the learning scheduler. The best performance for each dataset and backbone is highlighted in \textbf{bold}.}
\label{tab:soft-reg}
\renewcommand{\arraystretch}{1.2}
\setlength{\tabcolsep}{4pt}
\small
\begin{tabular}{l|cc|cc|cc|cc}
\hline
\multirow{2}{*}{Regularizer} 
& \multicolumn{2}{c|}{MMGCN} 
& \multicolumn{2}{c|}{DialogueGCN} 
& \multicolumn{2}{c|}{BiDDIN} 
& \multicolumn{2}{c}{MM-DFN} \\ 
& w-F1 & Acc & w-F1 & Acc & w-F1 & Acc & w-F1 & Acc \\ 
\hline
\multicolumn{9}{c}{\textit{IEMOCAP}} \\ 
\hline
Hard Regularizer 
    & \textbf{67.84} & \textbf{68.02} 
    & 66.46 & 66.61 
    & \textbf{59.98} & \textbf{60.07} 
    & \textbf{67.92} & \textbf{68.21} \\
Soft Linear 
    & 65.20 & 65.13 
    & 64.87 & 64.70 
    & 56.77 & 57.18 
    & 65.94 & 65.80 \\
Soft Logistic 
    & 65.98 & 66.17 
    & \textbf{67.14} & \textbf{67.41} 
    & 58.16 & 59.52 
    & 64.79 & 64.88 \\
\hline
\multicolumn{9}{c}{\textit{MELD}} \\ 
\hline
Hard Regularizer 
    & \textbf{59.35} & \textbf{61.72} 
    & 55.37 & 60.38 
    & \textbf{57.76} & \textbf{60.50} 
    & \textbf{59.14} & \textbf{62.07} \\
Soft Linear 
    & 58.43 & 60.73 
    & 54.55 & 60.12 
    & 57.64 & 59.62 
    & 57.64 & 59.62 \\
Soft Logistic 
    & 59.02 & 61.17 
    & \textbf{56.07} & \textbf{60.84} 
    & 56.94 & 59.72 
    & 58.27 & 61.26 \\
\hline
\end{tabular}
\end{table}

\begin{figure}[!ht]
    \centering

    \begin{minipage}[t]{0.95\textwidth}
        \centering
        \includegraphics[width=\textwidth]{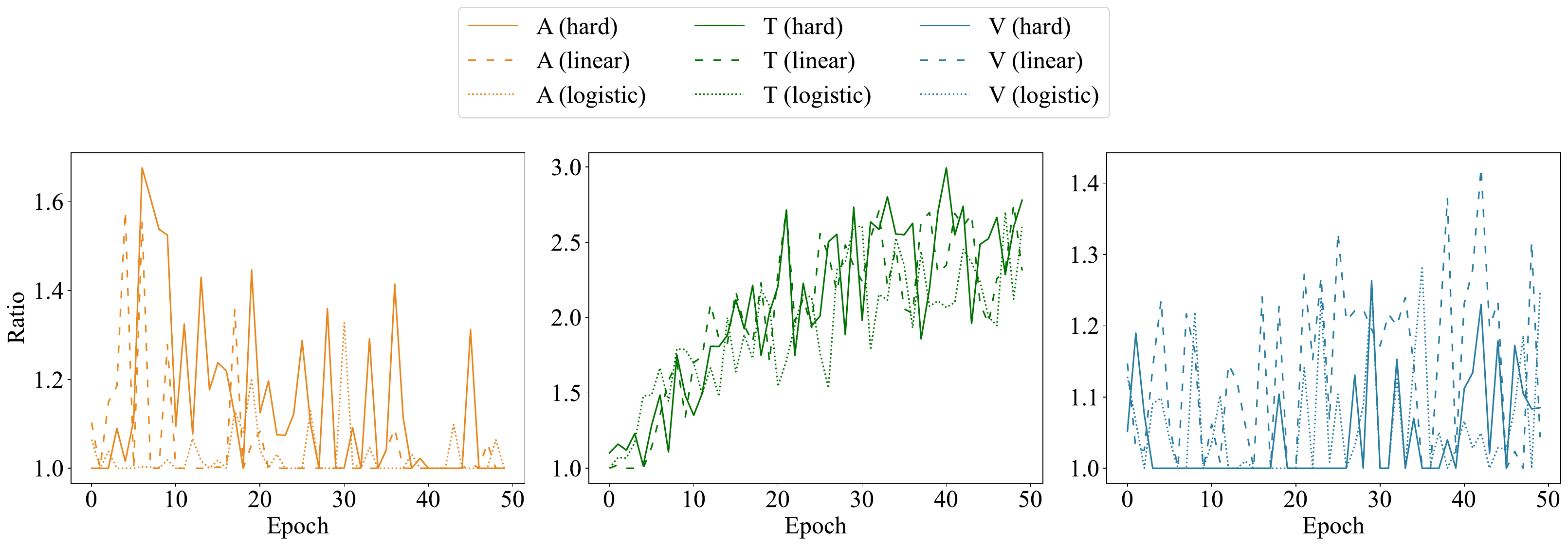}
        \par\small (a) DialogueGCN
    \end{minipage}
    
    \vskip1em

    \begin{minipage}[t]{0.95\textwidth}
        \centering
        \includegraphics[width=\textwidth]{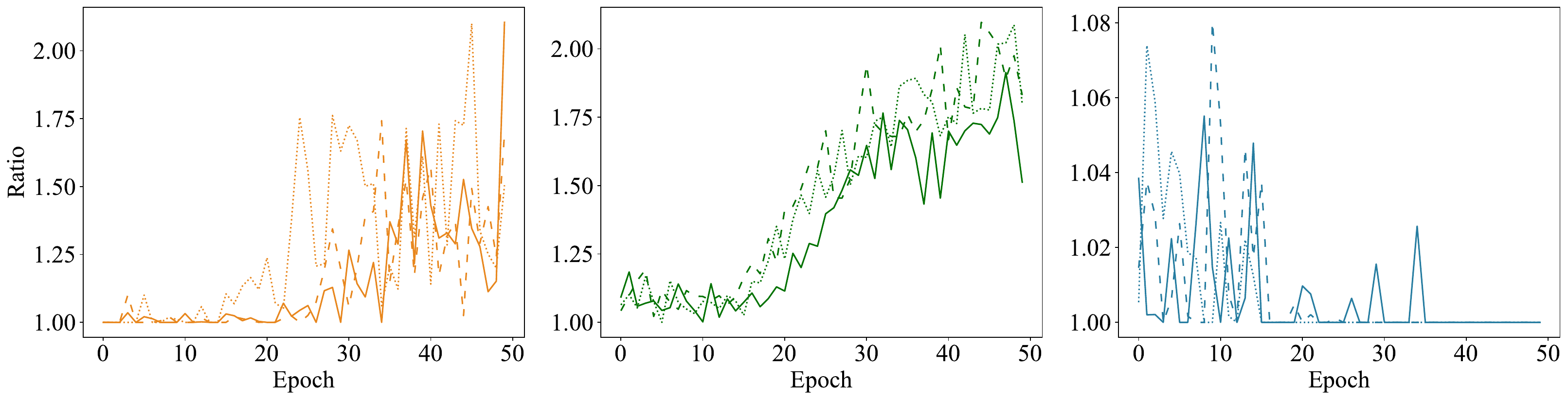}
        \par\small (b) BiDDIN
    \end{minipage}

    \vskip1em

    \begin{minipage}[t]{0.95\textwidth}
        \centering
        \includegraphics[width=\textwidth]{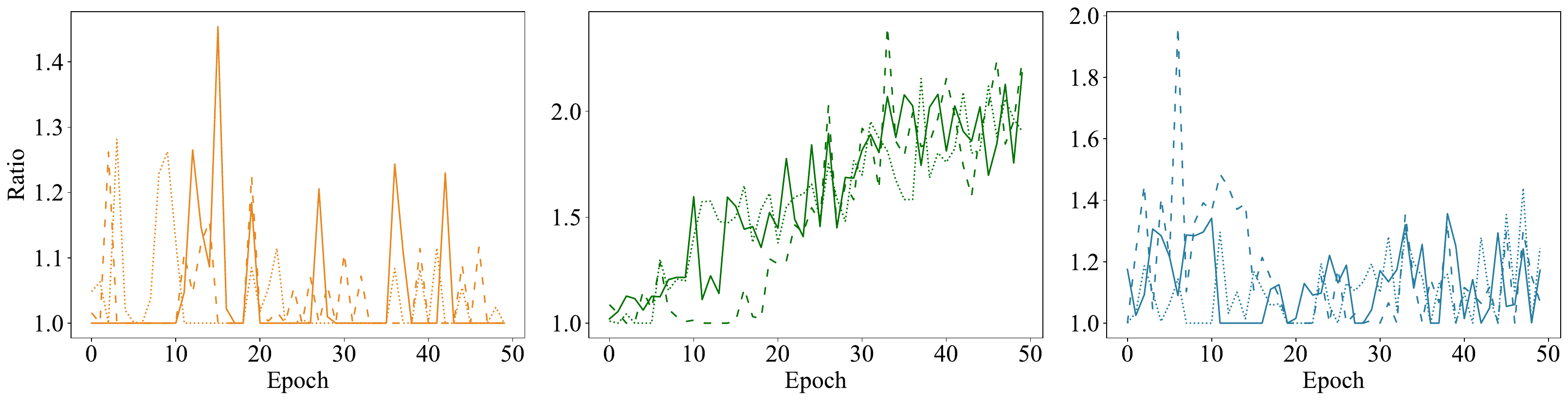}
        \par\small (c) MMGCN
    \end{minipage}

    \vskip1em

    \begin{minipage}[t]{0.95\textwidth}
        \centering
        \includegraphics[width=\textwidth]{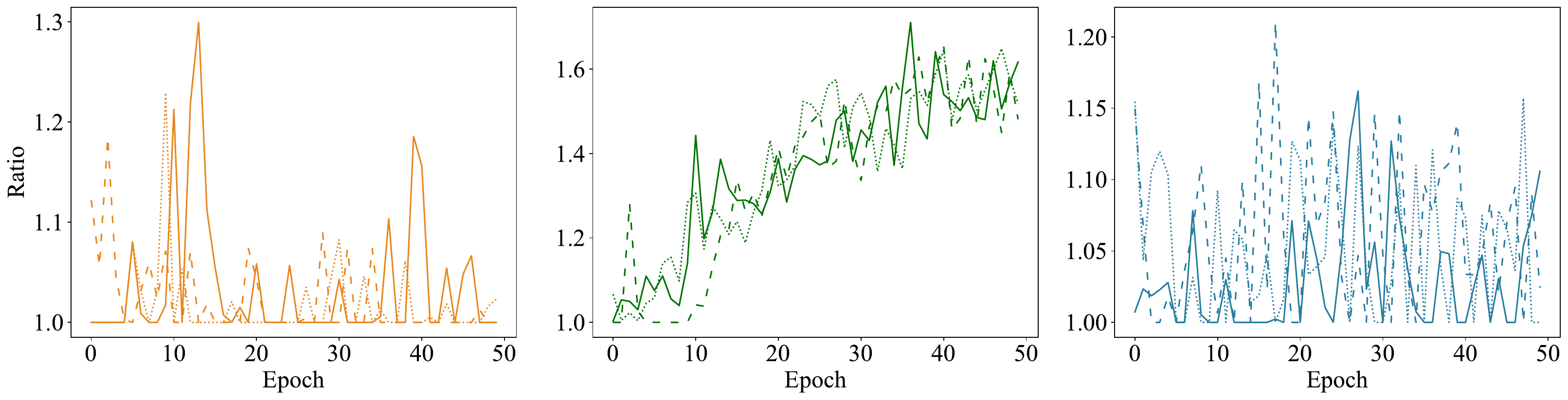}
        \par\small (d) MM-DFN
    \end{minipage}

    \caption{Modality ratio of the four backbones during training on the IEMOCAP dataset using different types of regularizer for the Learning Scheduler.}
    \label{fig:soft-reg}
\end{figure}

As shown in Table~\ref{tab:soft-reg} and illustrated in Figure~\ref{fig:soft-reg}, the hard regularizer consistently achieves superior or at least comparable performance across all evaluated backbones. For example, in the MMGCN model on the IEMOCAP dataset, the hard regularizer attains a weighted F1-score of 67.84\%, outperforming both the Linear (65.20\%) and Logistic (65.98\%) regularizers.
A similar trend is observed on the MELD dataset, where the hard regularizer achieves a weighted F1-score of 59.35\% in MMGCN, surpassing the Linear (58.43\%) and Logistic (59.02\%) alternatives.

Beyond accuracy, the hard regularizer also leads to more stable performance across modalities, contributing to reducing the discrepancy caused by modality imbalance. These findings confirm that a hard regularization strategy is more effective for our dual objectives: improving overall model performance and managing modality imbalance in multimodal emotion recognition.

\subsubsection{Analysis of Pacing Strategy}

We further study alternative strategies for updating the difficulty threshold $\lambda$ by adopting the following methods: cosine pacing, moving average(MA) pacing, and competence-based(CB) pacing~\cite{platanios2019competence}. The exponential pacing used in \mName{} is described in Eq.~\ref{eq:lambda-upd}, whereas the formulations of newly adopted strategies are described in Table~\ref{tab:pacing-equation}.

\begin{figure}
    \centering
    \includegraphics[width=0.9\linewidth]{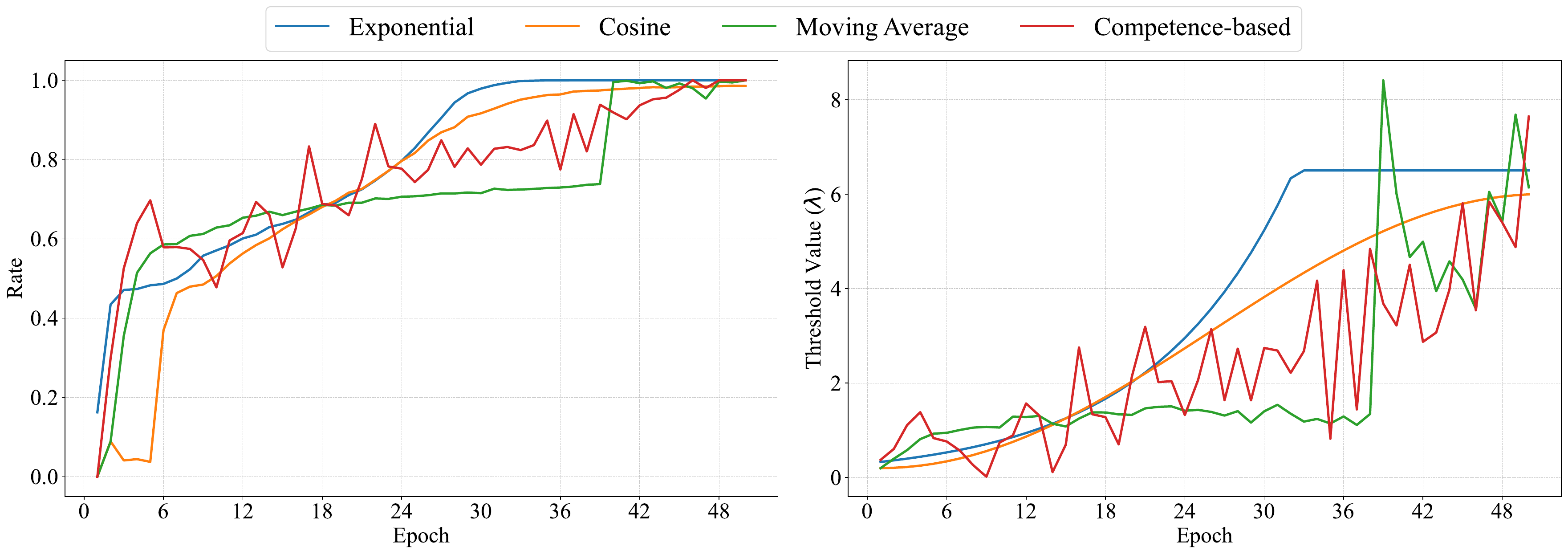}
    \caption{Curricula expanding rate and respective threshold value of MMGCN on MELD under different pacing strategies.}
    \label{fig:pacing}
\end{figure}

As shown in Fig.~\ref{fig:pacing}, linear pacing strategies, i.e., exponential and cosine pacing, yield smoother updates of $\lambda$. In contrast, the two non-linear strategies, where $\lambda$ is adaptively updated with regards to sample difficulty $\rho_{ij}$, exhibit larger fluctuations, particularly under competence-based pacing. Consequently, linear pacing results in a smoother curriculum expansion, indicating a more stable introduction of new samples. Table~\ref{tab:pacing} further shows that exponential and cosine pacing achieve better overall performance. These results highlight the importance of selecting an appropriate pacing strategy to ensure stable curriculum progression. 

\begin{table}[!ht]
    \centering
    \caption{Formulations of experimented pacing strategies. $T$ and $t$ denote total training epoch and current training epoch, respectively.}
    \label{tab:pacing-equation}
    \renewcommand{\arraystretch}{1.2} 
\setlength{\tabcolsep}{4.5pt} 
\small
\begin{tabular}{l|c}
        \hline
        \textbf{Strategy} & \textbf{Formulation} \\
        \midrule
        Cosine & $\lambda^{(t)}=\lambda_{\min}+\frac{\lambda_{\max}-\lambda_{\min}}{2}\cdot(1-\cos(\frac{\pi\cdot t}{T}))$ \\
        \midrule
        MA & $\lambda^{(t)}=\begin{cases}
            \alpha\lambda^{(t-1)}+(1-\alpha)\cdot\sum_i^{|\mathcal{D}|}\sum_j^{N_i}\rho_{ij} &\text{if }t<t_0\\
            \max\rho_{ij} &\text{if } t\geq t_0
        \end{cases}$ \\
        \midrule
        CB & $\begin{aligned}
c_t &= \min\!\left(1,\; \sqrt{ t \frac{1 - c_0^2}{T} + c_0^2 } \right) \\
\lambda^{(t)} &= \operatorname{Quantile}(\rho_{ij}, c_t)
\end{aligned}$ \\
\bottomrule
\end{tabular}
\end{table}

\begin{table}[!ht]
\centering
\caption{Performance comparison of MMGCN and MM-DFN on IEMOCAP and MELD dataset using different pacing strategies. The best and second-best performances for each dataset and backbone are highlighted in \textbf{bold} and \uline{underline}.}
\label{tab:pacing}
\renewcommand{\arraystretch}{1.2}
\setlength{\tabcolsep}{4pt}
\small
\begin{tabular}{l|cc|cc|cc|cc}
\toprule
\multirow{3}{*}{Strategy} 
& \multicolumn{4}{c|}{\textit{IEMOCAP}}
& \multicolumn{4}{c}{\textit{MELD}} \\
\cmidrule(lr){2-5} \cmidrule(lr){6-9}
& \multicolumn{2}{c|}{MMGCN} 
& \multicolumn{2}{c|}{MM-DFN} 
& \multicolumn{2}{c|}{MMGCN} 
& \multicolumn{2}{c}{MM-DFN} \\ 
& w-F1 & Acc & w-F1 & Acc & w-F1 & Acc & w-F1 & Acc \\ 
\hline
Exponential & \textbf{67.84} & \textbf{68.02} & \textbf{67.92} & \textbf{68.21} & \textbf{59.35} & \uline{61.72} & \textbf{59.14} & \textbf{62.07} \\
Cosine      & \uline{65.47} & \uline{65.00} & \uline{67.63} & \uline{67.53} & 58.10 & 61.23 & 57.06 & \uline{61.11} \\
MA          & 62.63 & 62.91 & 67.55 & 67.34 & 57.78 & 60.77 & 56.46 & 57.78 \\
CB          & 63.80 & 63.52 & 66.68 & 66.42 & \uline{58.71} & \textbf{62.34} & \uline{58.44} & 60.61 \\
\hline
\end{tabular}
\end{table}

\begin{figure}[!ht]
    \centering

    \begin{minipage}[t]{\textwidth}
        \centering
        \includegraphics[width=\textwidth]{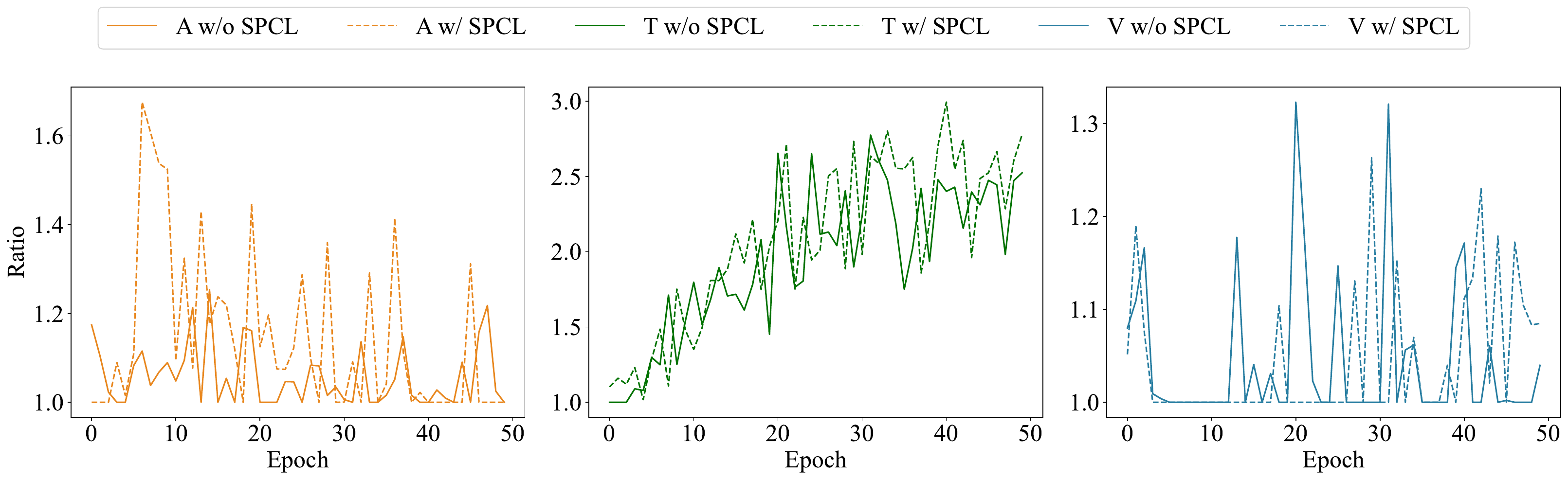}
        \par\small (a) DialogueGCN
    \end{minipage}
    
    \vskip\baselineskip

    \begin{minipage}[t]{\textwidth}
        \centering
        \includegraphics[width=\textwidth]{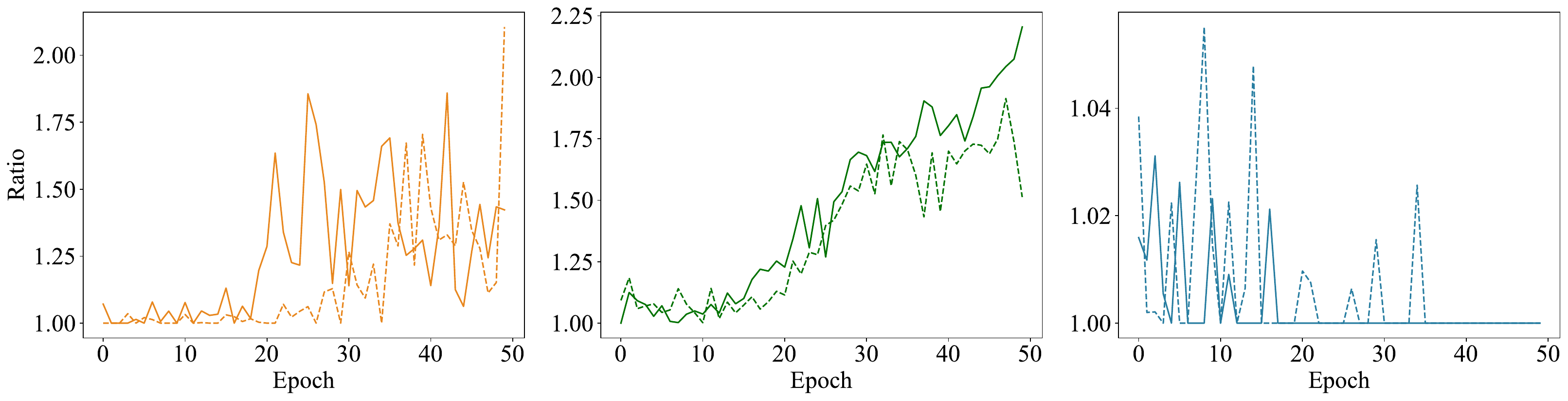}
        \par\small (b) BiDDIN
    \end{minipage}
    
    \vskip\baselineskip

    \begin{minipage}[t]{\textwidth}
        \centering
        \includegraphics[width=\textwidth]{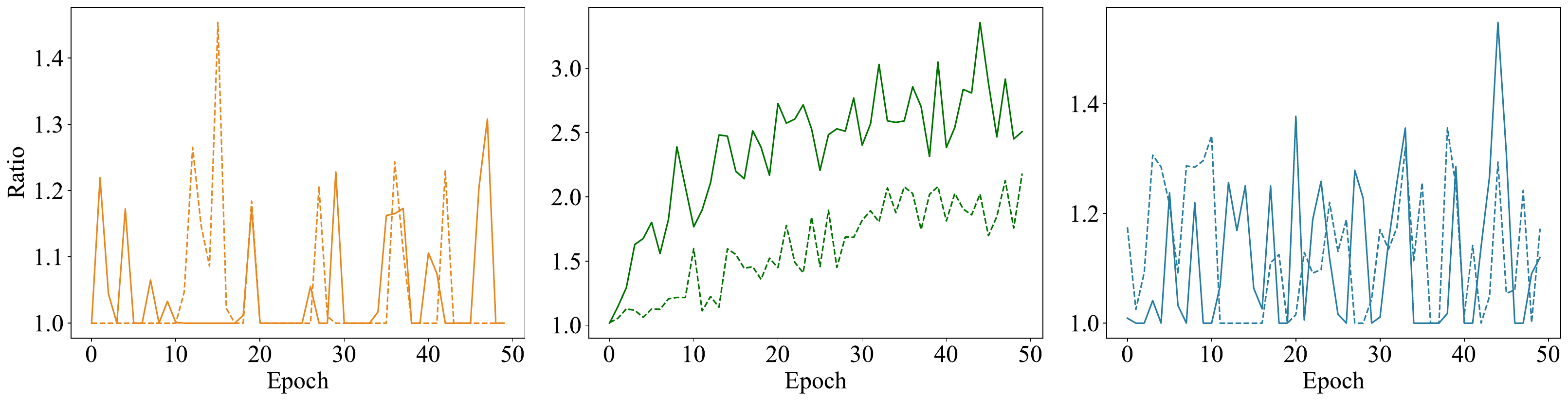}
        \par\small (c) MMGCN
    \end{minipage}
    
    \vskip\baselineskip

    \begin{minipage}[t]{\textwidth}
        \centering
        \includegraphics[width=\textwidth]{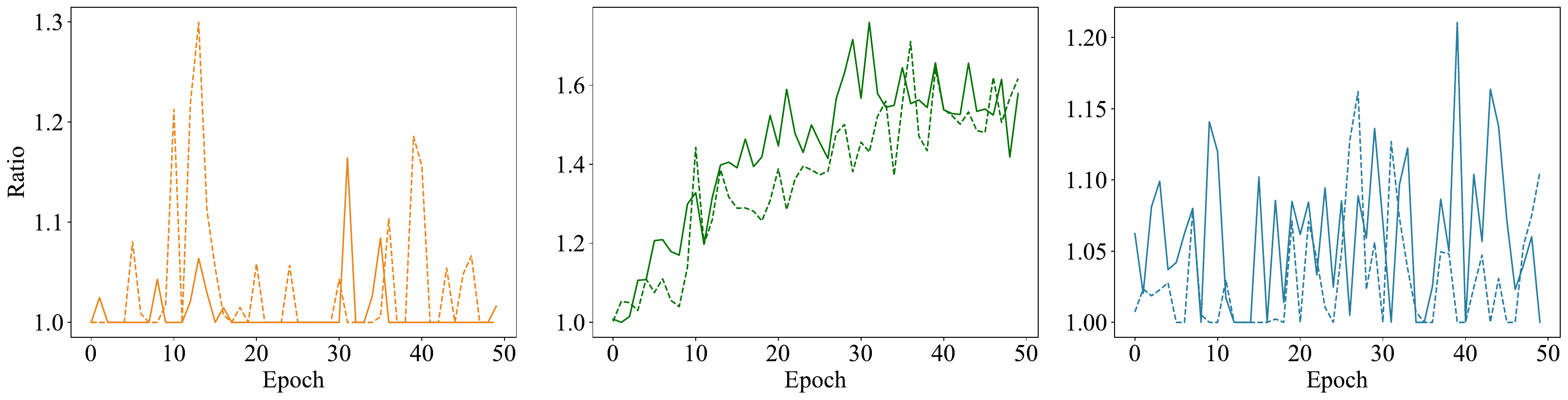}
        \par\small (d) MM-DFN
    \end{minipage}

    \caption{Modality ratio of the four backbones during training on the IEMOCAP dataset.}
    \label{fig:modality-ratio}
\end{figure}
\subsubsection{Modality Ratio}
Our study aims to achieve two key objectives: \textit{(1) enhancing the performance of tri-modal models relative to their bi-modal and uni-modal counterparts and (2) mitigating modality imbalance during training}. To further investigate the latter, we analyze the modality ratio, which quantifies each modality’s contribution relative to the weakest modality throughout training. 

As depicted in Figure \ref{fig:modality-ratio}, the integration of our \mName{} module effectively reinforces the weaker modalities across all baseline models. Specifically, we observe an increase in the audio modality ratio by 0.2 to 0.5 and an increase of 0.15 in the visual modality ratio. Concurrently, our approach reduces the dominance of the strongest modality (i.e. text). This effect is particularly notable in MMGCN, where the text modality ratio decreases from 3 to 2, indicating a more balanced learning process. These findings confirm that our method successfully addresses modality imbalance by narrowing the gap between strong and weak modalities, ensuring a more equitable contribution from all modalities. 
\subsubsection{Limitations}

Although SPCL introduces negligible architectural overhead, several practical considerations remain. First, its effectiveness is sensitive to backbone architectures and dataset characteristics, often requiring extensive hyperparameter tuning. As discussed in Section~\ref{sec:tuning}, brute-force strategies such as grid search are impractical for large-scale models, underscoring the need for a more general and adaptive tuning strategy.
Second, the Difficulty Measurer operates at the batch level rather than over the entire dataset, which may lead to inaccurate difficulty estimation when easy or hard samples are unevenly distributed. Increasing batch size can mitigate this, though it also raises computational demands,  highlighting a trade-off between estimation reliability and resource efficiency.

From a scalability perspective, SPCL was designed as a plug-in module with minimal computational overhead. At each training step, the Difficulty Measurer reuses model outputs and losses to compute utterance- and conversation-level scores, while the Learning Scheduler adjusts sample weights without redundant computation. With efficient masking and matrix operations implemented in PyTorch and NumPy, the additional training time remains moderate (e.g., an average increase of around 10s per epoch for MMGCN on MELD compared to IEMOCAP).

Data-wise, SPCL’s curriculum progression may not always align with datasets containing brief or fragmented dialogues. Moreover, both benchmark datasets (IEMOCAP and MELD) consist of scripted dialogues, which may not fully capture the dynamics of spontaneous emotional interactions. Future work will extend SPCL to more naturalistic datasets (e.g., K‑EmoCon~\cite{park2020k}) to evaluate its robustness and generalizability in real-world scenarios.

Finally, since SPCL operates solely during the training phase and leaves the baseline model architecture unchanged, it can be seamlessly integrated into various multimodal frameworks and does not affect model deployment, further demonstrating its extensibility and practicality for large-scale conversational emotion recognition.

\section{Conclusion}
\label{sec:conclude}
In this work, we have introduced \mName{}, a plug-and-play module designed to address modality imbalance in Multimodal Emotion Recognition in Conversation (MERC). Our approach leverages Self-Paced Curriculum Learning to dynamically mitigate modality discrepancies during training, thereby promoting more balanced multimodal representation learning. Specifically, \mName{} comprises two key components: (1) a Difficulty Measurer, which quantifies sample complexity at both the utterance and conversation levels based on loss dynamics and modality alignment, and (2) a Learning Scheduler, which adaptively regulates the training curriculum, progressing from easier to more complex samples.
Extensive experiments on IEMOCAP and MELD validate the effectiveness of \mName{}, demonstrating consistent improvements in both w-F1 and accuracy across multiple baselines. Moreover, our approach effectively strengthens weaker modalities while mitigating the over-reliance on dominant ones, leading to a more balanced multimodal representation. Ablation studies further highlight the necessity of integrating both utterance- and conversation-level information and maintaining a progressive curriculum expansion rate for optimal convergence.

We further show that maintaining an appropriate curriculum expanding rate is essential. A rate that grows steadily over training yields better results, as it allows weaker modalities sufficient exposure to easier samples while preventing overfitting on simple instances. Additionally, our comparison between regularization strategies demonstrates that the hard regularizer consistently achieves superior or comparable performance to soft regularizers, offering more stable learning across modalities and better managing modality imbalance. Moreover, our analysis of modality ratios confirms that \mName{} effectively reduces the dominance of strong modalities, such as text, and enhances contributions from weaker ones like audio and visual inputs, thereby addressing modality imbalance more comprehensively.

In future work, we aim to enhance our approach by adapting the curriculum scheduling strategy to better reflect dataset-specific traits, exploring alternative difficulty measures suited to brief dialogues, and integrating adaptive pacing mechanisms to ensure greater compatibility with diverse conversational formats. Further extensions include expanding SPCL for Large Multimodal Model (LMMs), and to cross-lingual datasets. To support these developments, comprehensive studies on hyperparameter behaviors and evaluations on more diverse datasets will be conducted to address current limitations and ensure robust scalability. Additionally, we find extending SPCL to the task of sentiment analysis in conversations presents a feasible and meaningful direction, given its close similarity in data characteristics and backbone architectures to multimodal emotion recognition. 

\section*{Declaration of Competing Interests}
The authors declare that they have no known competing financial interests or personal relationships that could have appeared to influence this work.

\section*{Data Availability Statement}
The experimental data in this study are from the IEMOCAP (\hyperlink{https://sail.usc.edu/iemocap/iemocap.htm}{https://sail.usc.edu/iemocap/iemocap.htm}) and MELD (\hyperlink{https://affective-meld.github.io/}{https://affective-meld.github.io/}). 

\section*{Code availability}
The code will be made available upon publications.

\bibliography{_manuscript}

\end{document}